%% file: main.tex
\documentclass{article}

\usepackage{microtype}
\usepackage{graphicx}
\usepackage{subfigure}
\usepackage{booktabs} % for professional tables

\usepackage[utf8]{inputenc} % allow utf-8 input
\usepackage[T1]{fontenc}    % use 8-bit T1 fonts
\usepackage{url}            % simple URL typesetting
\usepackage{amsfonts}       % blackboard math symbols
\usepackage{nicefrac}       % compact symbols for 1/2, etc.
\usepackage{xcolor}         % colors
\usepackage{amsmath,amssymb,amsfonts}

\usepackage{multirow}
\usepackage{multicol}
\usepackage{algorithm}
\usepackage{algorithmic}
\usepackage{wrapfig}

\usepackage{framed}
\usepackage{listings}
\usepackage{xcolor}
\usepackage{xspace}
\definecolor{codegreen}{rgb}{0,0.6,0}
\definecolor{codegray}{rgb}{0.5,0.5,0.5}
\definecolor{codepurple}{rgb}{0.58,0,0.82}
\definecolor{backcolour}{rgb}{0.95,0.95,0.92}

\usepackage{hyperref}

\newcommand{\trim}{\vspace{-3mm}}

\lstdefinestyle{mystyle}{
    backgroundcolor=\color{backcolour},
    commentstyle=\color{codegreen},
    keywordstyle=\color{magenta},
    morekeywords={"self"},
    numberstyle=\tiny\color{codegray},
    stringstyle=\color{codepurple},
    basicstyle=\ttfamily\scriptsize,
    breakatwhitespace=false,         
    breaklines=false,
    captionpos=b,            
    keepspaces=true,                 
    numbers=left,                    
    numbersep=3pt,                  
    showspaces=false,                
    showstringspaces=false,
    showtabs=Tr,                  
    tabsize=2,
    frame=single, 
    framerule=0pt
}
\usepackage[accepted]{mlsys2022}

\mlsystitlerunning{DGI: Easy and Efficient Inference for GNNs}

\newtheorem{layer-wise}{Property}
\newcommand{\sys}{DGI\xspace}

\newcommand{\stitle}[1]{\vspace*{0.4em}\noindent{\bf #1.\/}}

\begin{document}

\twocolumn[
\mlsystitle{DGI: Easy and Efficient Inference for GNNs}

% unless you've provided the [accepted] option to the mlsys2022


\mlsyssetsymbol{equal}{*}

\begin{mlsysauthorlist}
\mlsysauthor{Peiqi Yin}{aws,cuhk,sustech}
\mlsysauthor{Xiao Yan}{sustech}
\mlsysauthor{Jinjing Zhou}{tc}
\mlsysauthor{Qiang Fu}{gwu}
\mlsysauthor{Zhenkun Cai}{aws}
\mlsysauthor{James Cheng}{cuhk}
\mlsysauthor{Bo Tang}{sustech}
\mlsysauthor{Minjie Wang}{aws}
\end{mlsysauthorlist}

\mlsysaffiliation{sustech}{Southern University of Science and Technology}
\mlsysaffiliation{aws}{AWS Shanghai AI Lab}
\mlsysaffiliation{tc}{TensorChord}
\mlsysaffiliation{gwu}{The George Washington University}
\mlsysaffiliation{cuhk}{The Chinese University of Hong Kong}

\mlsyscorrespondingauthor{Minjie Wang}{minjiw@amazon.com}

\mlsyskeywords{Machine Learning, MLSys}

\vskip 0.3in

\begin{abstract}
%While many systems for training Graph Neural Networks (GNNs) have been proposed in recent years, scalable model inference and evaluation remain to be inadequately studied. Model evaluation using the widely adopted \textit{node-wise} approach can account for up to 94\% of the time in the end-to-end training process due to the neighbor explosion issue. Moreover, \textit{node-wise} inference requires the entire graph to fit into CPU memory, thus making it difficult for commodity hardware with limited memory capacity. In this paper, we formalize an efficient \textit{layer-wise} inference approach which is suitable for out-of-core execution. In addition, to mitigate the programming difficulty of \textit{layer-wise} inference, we develop \sys~--- a easy and efficient system that automatically translates a user-defined GNN model for layer-wise execution. \sys further accelerates model inference by node reordering and dynamic batch size control. Our experimental results show that \sys can speed up model inference by up to three orders of magnitude and run on machines with small memory capacity.

% Moreover, \textit{node-wise} inference requires the entire graph to fit into CPU memory, thus making it difficult for commodity hardware with limited memory capacity.
While many systems have been developed to train Graph Neural Networks (GNNs), efficient model inference and evaluation remain to be addressed. For instance, using the widely adopted \textit{node-wise} approach, model evaluation can account for up to 94\% of the time in the end-to-end training process due to \textit{neighbor explosion}, which means that a node accesses its multi-hop neighbors. On the other hand, \textit{layer-wise} inference avoids the neighbor explosion problem by conducting inference layer by layer such that the nodes only need their one-hop neighbors in each layer. However, implementing layer-wise inference requires substantial engineering efforts because users need to manually decompose a GNN model into layers for computation and split workload into batches to fit into device memory. In this paper, we develop \textit{Deep Graph Inference (\sys)} --- a system for easy and efficient GNN model inference, which automatically translates the training code of a GNN model for layer-wise execution. \sys is general for various GNN models and different kinds of inference requests, and supports out-of-core execution on large graphs that cannot fit in CPU memory. Experimental results show that \sys consistently outperforms layer-wise inference across different datasets and hardware settings, and the speedup can be over 1,000x.

%for different GNN models and inference requests

%and run on machines with small memory capacity
%For the easy adoption of layer-wise inference

\end{abstract}
]



\printAffiliationsAndNotice{}  % leave blank if no need to mention equal contribution
% \printAffiliationsAndNotice{\mlsysEqualContribution} % otherwise use the standard text.

\input{sections/1_introduction.tex}

\input{sections/2_layerwise.tex}

\input{sections/3_methodology.tex}
\input{sections/4_optimizations.tex}
\input{sections/5_evaluation.tex}

\input{sections/6_related_works.tex}

\input{sections/7_conclusion.tex}

\clearpage

\appendix
\section{An example of hand-written layer-wise inference for JKNet}

\begin{lstlisting}[language=Python, caption=Simplified handwritten code for the layer-wise inference of JKNet., label=code:manual]
# We adopt the forward function of JKNet from the DGL
# library, and implement layer-wise inference for JKNet
# manually.
class JKNet(nn.Module):
    def __init__(...):
        ... # init for JKNet.

    def forward(self, g, x):
        hs = []
        h = x
        for layer in self.layers:
            h = layer(g, h)
            h = self.dropout(h)
            hs.append(h)
        h = torch.cat(hs, dim=-1)
        return self.conv(h)

    def inference(self, g, batch_sizes, x):
        feat_lst = []
        for l, layer in enumerate(self.layers):
            ret = torch.zeros(
                (g.num_nodes(), self.hidden))
            feat_lst.append(ret)
            dl = NodeDataLoader(
                batch_size=batch_sizes[0], ...)
            for in_nodes, out_nodes, blocks in dataloader:
                block = blocks[0].to("GPU")
                h = x[in_nodes].to("GPU")
                h = layer(block, h)
                h = self.dropout(h)
                feat_lst[-1][out_nodes] =  h.cpu()
            x = feat_lst[-1]

        ret = torch.zeros(
            (g.num_nodes(), self.n_classes))
        dl = NodeDataLoader(
                batch_size=batch_sizes[1], ...)
        for in_nodes, out_nodes, blocks in dataloader:
            block = blocks[0].to("GPU")
            h_lst = []
            for feat in feat_lst:
                h_lst.append(feat[in_nodes].to("GPU"))
                h = torch.cat(h_lst, dim=-1)
                out_val = self.conv(h)
            ret[out_nodes] = out_val.cpu()
        return ret
\end{lstlisting}

\begin{lstlisting}[language=Python, caption=Code for using DGI for layer-wise inference of JKNet., label=code:dgi]
graph = ... # Graph data
x = ... # Node features
nids = torch.arange(graph.num_nodes())
model = JKNet(...)

# Inference by DGI.
from dgl.dataloading import FullNeighborSampler
from dgi import AutoInferencer
infer = AutoInferencer(model,
            targets=nids,
            sampler=FullNeighborSampler(1),
            computing_device="GPU:0",
            external_device="CPU")
# Here takes the same arguments as 
# forward function in JKNet.
pred = infer.inference(graph, x)
\end{lstlisting}

To demonstrate the challenges of coding layer-wise inference and the usability of \sys, Listing~\ref{code:manual} shows how to use handwritten code to conduct layer-wise inference for JKNet. 
For the forward function, JKNet first computes each layer's embeddings and records them in an array. Finally, it concretes all embeddings and uses an additional graph convolution to perform the output. 
It consists of only 8 LOCs (i.e., Line 8-16 in Listing~\ref{code:manual}) and node-wise influence is easy to implement by directly using the forward function.
However, layer-wise inference needs approximately 30 LOCs (from Line 18 in  Listing~\ref{code:manual}) even the code is already simplified by removing some of the complex and troublesome API calls. 
Line 20 first enumerate convolution layers, and perform layer-wise inference. 
For the additional graph convolution layer in JKNet, we need to write another layer-wise code for it (line 38 - 45). 
Note that the final layer of JKNet (i.e., the jumping knowledge layer) aggregates the output embedding of all preceding layers. 
The handwritten code needs to explicitly consider whether UVA memory is used in the CPU and manually configure the batch size for the dataloader. 
If data are stored on SSD, the handwritten code will be even more complex.
Besides, the inference speed of layer-wise execution is sensitive to multiple factors such as the size of each node batch and the order to process nodes. 

Listing~\ref{code:dgi} shows how to perform layer-wise inference using DGI. DGI provide interface that users only need to input the model to initialize an \textit{AutoInferencer}, which will automatically trace the split the forward function and pack to the inference function. Then, users only need to call the inference function provide by \textit{AutoInferencer}, which takes the arguments as the same as the model forward function. DGI completely frees the user from having to write all this tedious code.

\begin{figure}[htbp]
		\centering
	\includegraphics[width=\linewidth,scale=0.9]{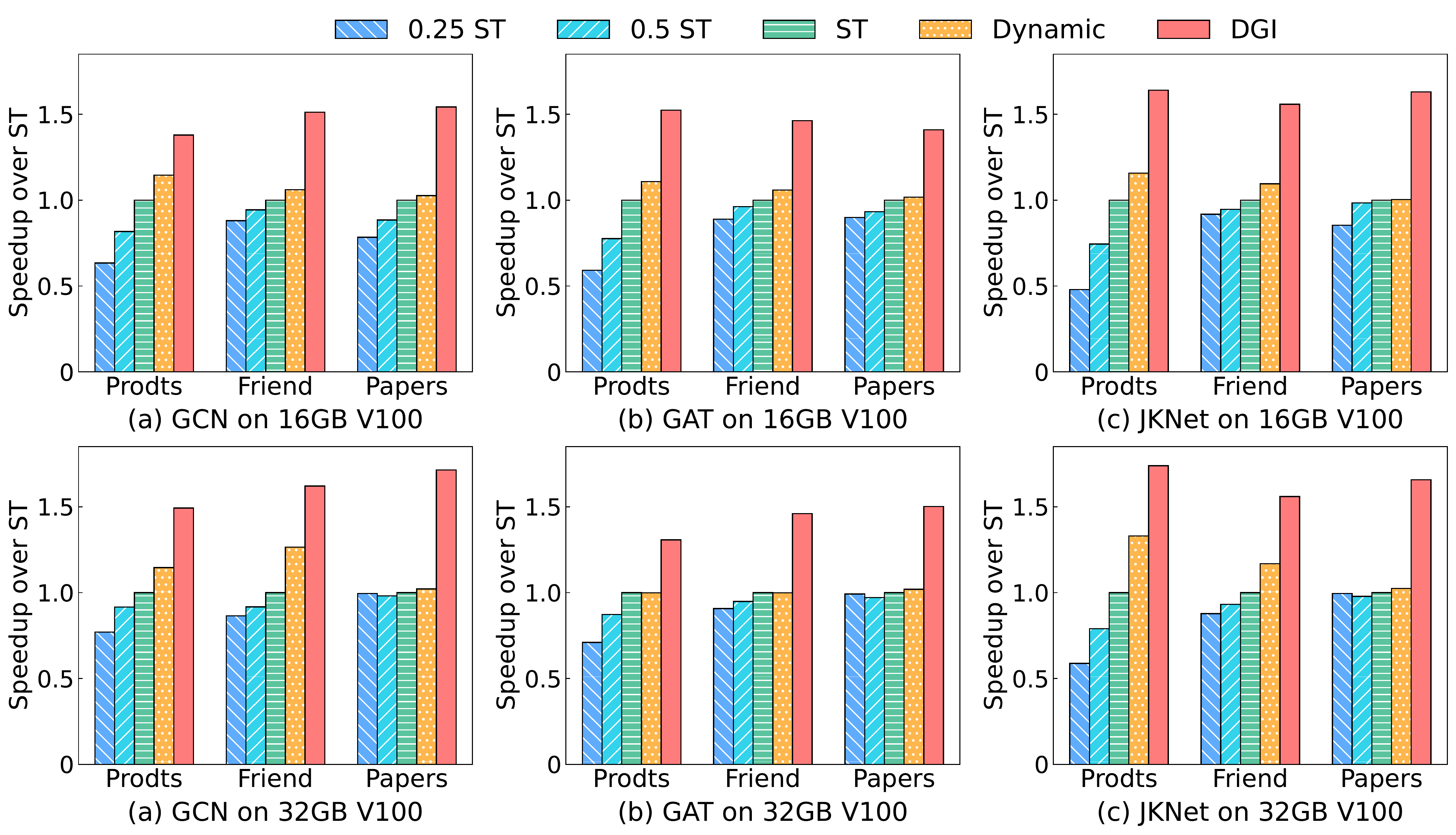}
  \caption{Speedup of dynamic batch size and node ordering in DGI over using the best static batch size found by profiling.
  %We report the inference speedup over ST baseline Comparison between the inference execution time of \sys and the baselines. As the execution time for different graphs varies in scale, we normalize the execution time w.r.t. ST. The numbers under each dataset are the inference execution time of \sys (in seconds).
  }
  \label{fig:ablation_study}
\end{figure}

\section{More Experiments Comparing the DGI Variants}

We compare the DGI variants (introduced in Figure 8 of the main paper) on the V100 GPU with 32GB memory in Figure~\ref{fig:ablation_study}. To observe the influence of GPU memory, we also include the results on the V100 GPU with 16GB memory alongside. The results show that dynamic batch size and node ordering are effective for both GPU configurations---\textit{Dynamic} performs close to or better than \textit{ST}, and DGI significantly outperforms \textit{Dynamic} with node ordering. The performance gap among different batch sizes (i.e., \textit{0.25ST}, \textit{0.5ST} and \textit{ST}) is smaller for 32GB GPU memory than 16GB GPU memory because 32GB memory allows larger batch sizes for all baselines and the gains of input sharing diminish with batch size.
\nocite{langley00}

\bibliography{main}
\bibliographystyle{mlsys2022}

\end{document}

%% file: sections/1_introduction.tex
\section{Introduction}

% introduction to GNN

Graph data are ubiquitous in domains such as social networks~\citep{social1, social2}, knowledge representations~\citep{knowledge1}, and bio-informatics~\citep{bio1, bio2}. Recently, graph neural networks (GNNs) have shown outstanding performance for many tasks including node classification~\citep{nodeclassification1, ogbn}, clustering~\citep{clustering1}, and link prediction~\citep{linkpred1, linkpred2}. A plethora of GNN models with different structures have been proposed, e.g., GCN~\citep{GCN}, GAT~\citep{GAT}, JKNet~\citep{JKNet} and APPNP~\citep{APPNP}. These models generally stack $L$ graph aggregation (also called convolution) layers to compute an output embedding $h_v^L$ for each node $v$ in a graph $G \!=\! (V, E)$, and each layer can be expressed as       
\begin{equation}\label{equ:GNN}
	h_{v}^l = \text{AGGREGATE}^l({h_u^{l-1}, \forall u\in \mathcal{N}(v)}\cup v;w^l),
\end{equation}
where set $\mathcal{N}(v)$ contains the in-neighbors of node $v$, $h_v^l$ is the embedding of node $v$ in the $l\textsuperscript{th}$ layer (with $h_v^0$ being the input node embedding), and $w^l$ is the parameter of function $\text{AGGREGATE}^l$.

\begin{figure}[htbp]
	\centering
	\includegraphics[width=\linewidth,scale=0.9]{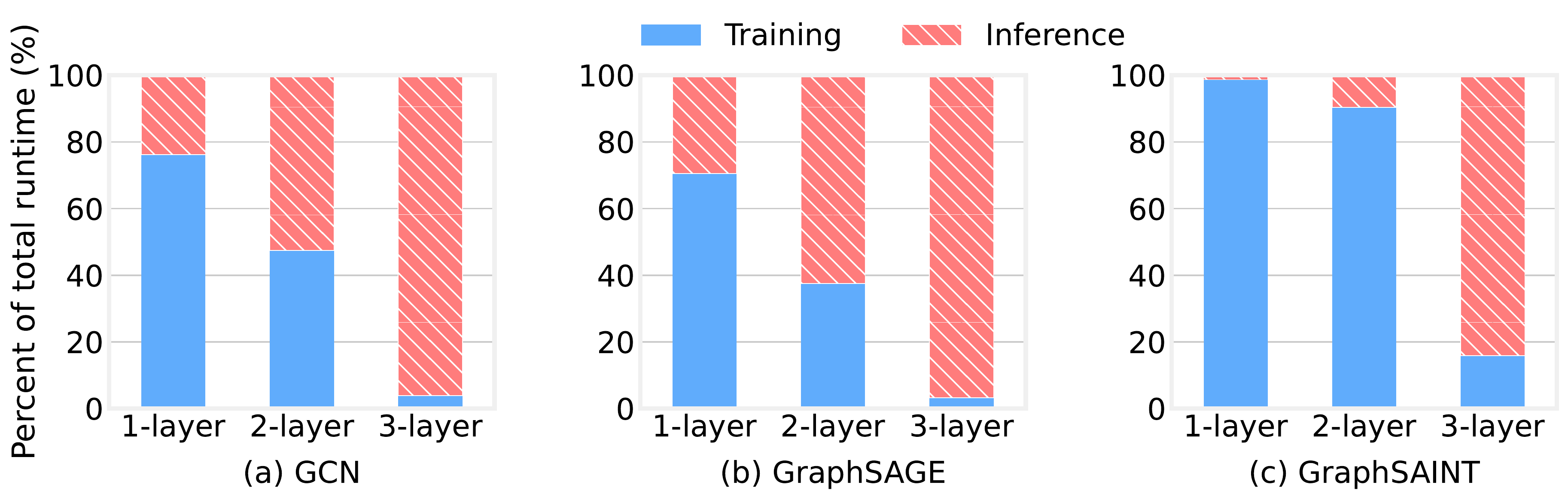}
	\caption{Composition of model training time and node-wise inference time in a model training pipeline for the \textit{OGBN-Products} graph on a V100 GPU with 32GB memory. All experiments conduct evaluation every 10 training epochs following the setting of models in OGB Leaderboard~\citep{ogbn}.}
	\label{fig:train_vs_infer}
	\trim
	%\trim
\end{figure}

%on the \textit{OGBN-Products} graph~\citep{ogbn}
GNN inference computes output embeddings for all (or a set of) nodes in the graph to facilitate model evaluation or downstream applications (e.g., recommendation). A natural solution is to utilize the code for training, which usually computes output embedding for each target node $v$ individually by first collecting its $L$-hop neighbors and then applying Eq.~\eqref{equ:GNN}. We call this approach \textit{node-wise inference} and report the time taken by model training and inference (evaluation) for some popular GNNs in Figure~\ref{fig:train_vs_infer}. The results show that inference time increases quickly with the number of layers and accounts for up to 94\% of the end-to-end time. This is because node-wise inference suffers from \textit{neighbor explosion}: 
(i) The $L$-hop neighbors of a target node and the input embeddings of these neighbors are enormous, which make data
preparation expensive. (ii) To fit the memory of GPU, inference is normally conducted in small batch. 
%(ii) The $L$-hop neighbors of the target nodes from different batches are overlapped, resulting repetitive feature loading. 
Target nodes in different batches compute intermediate embeddings (i.e., in layers with $l<L$) for their common neighbors, resulting in repetitive computation.
%(i) each target node collects its $L$-hop neTo fit the memory of GPU device, the inference normally is conducted in small batch. ighbors in the graph and the input embeddings of these neighbors, which results in high data preparation cost; (ii) different target nodes compute intermediate embeddings (i.e., in layers with $l<L$) for their common neighbors, yielding repetitive computation; (iii) to fit in memory of a computing device (e.g., GPU), inference is conducted in batches for the target nodes but a large input set (i.e., $L$-hop neighbors) for each node results in a small batch size, which makes it difficult to share inputs within a batch.
The neighbor explosion problem is prominent for inference because (i) it usually uses the full neighborhood as opposed to the neighbor sampling in training~\cite{SAGE,fastgcn,clustergcn}, and (ii) real-world graphs contain popular nodes with a large number of neighbors, and thus the size of $L$-hop neighborhood could be very large~\cite{ogbn, snap}.     

\begin{figure}[htbp]
	\centering
	\includegraphics[width=\linewidth,scale=0.9]{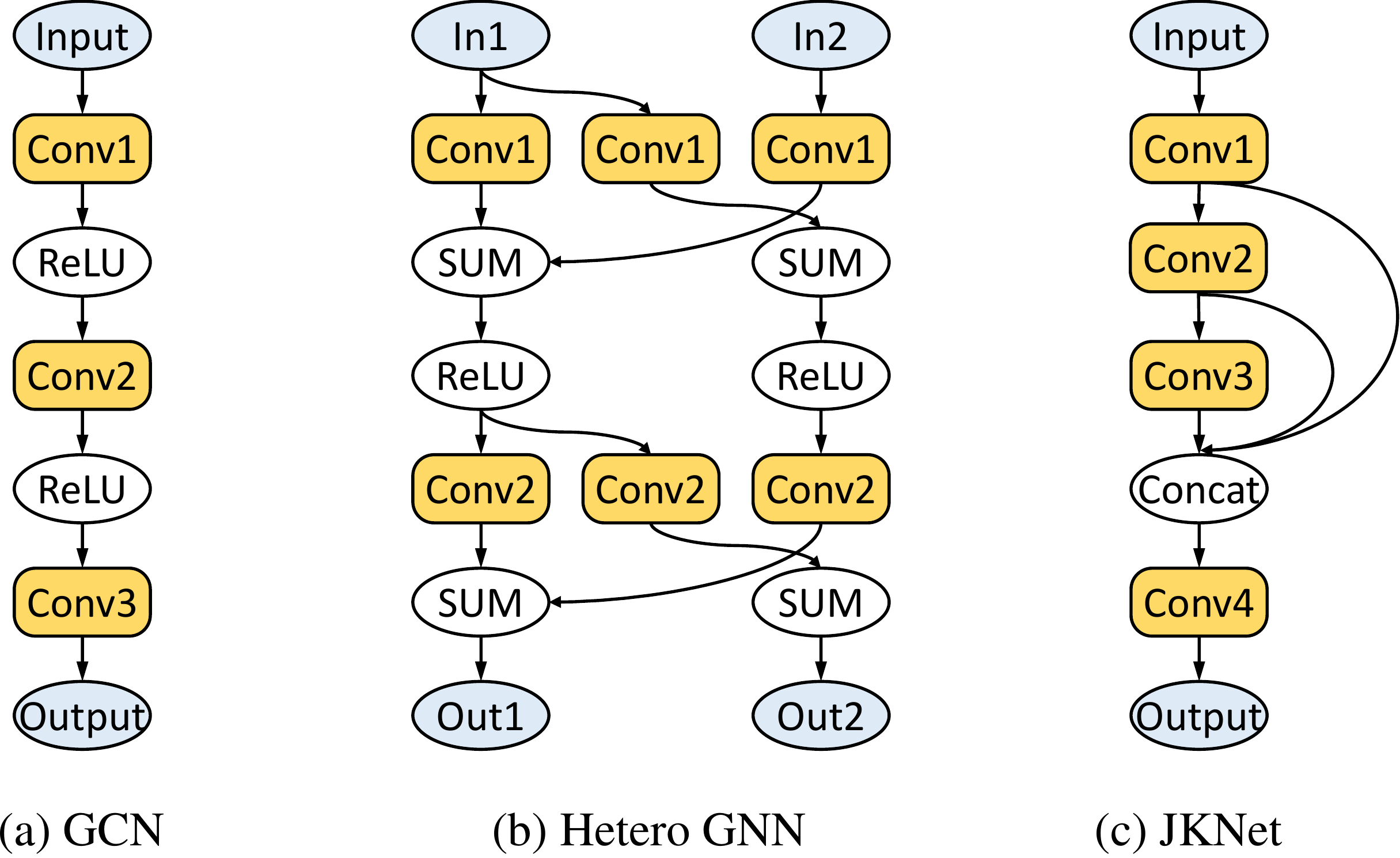}
	\caption{The structure of different GNN models. $\mathsf{Conv}$ is a graph aggregation layer, and we omit the activation functions in (c).}
	\label{fig:GNN structure}
\end{figure}

We present an efficient \textit{layer-wise inference} approach, which conducts computation layer by layer and handles the tasks of all target nodes in the same layer batch by batch. Layer-wise inference avoids the neighbor explosion problem as a node only accesses its 1-hop neighbors in each layer. The small input set also enables efficient out-of-core execution for large graphs by storing data in external memory and loading only the input set of each batch. Moreover, common computation tasks in a layer from different target nodes are merged to eliminate repetitive computation. However, substantial implementation efforts are required to enjoy the efficiency of layer-wise inference. In particular, users need to manually decompose a GNN model into layers, merge the computation tasks in each layer, and manage the intermediate embeddings and node batching. Figures~\ref{fig:GNN structure}(b-c) show that GNN models can be much more complex than the linear structure in Figure~\ref{fig:GNN structure}(a), and thus it is difficult to decompose them into layers that are suitable for inference. Moreover, the batch size is cumbersome to set in order to fully utilize memory while avoiding out-of-memory (OOM).

To mitigate the programming difficulty of layer-wise inference, we build \textit{Deep Graph Inference (DGI)} --- a system for easy and efficient GNN inference. \sys generalizes across GNN models (e.g., homogeneous and heterogeneous ones) and automatically translates the training code for layer-wise execution, and thus removes the burden of writing separate code for inference. In particular, DGI uses a tracer to resolve the computation graph of a GNN model, designs tailored rules to partition the computation graph into layers, and schedules the layers for computation. DGI also dynamically adjusts the batch size according to runtime statistics and reorders the graph nodes to form batches for better input sharing in each batch. DGI can handle different kinds of inference requests: (i) \textit{full inference}, which computes output embedding for all nodes in the graph, (ii) \textit{partial inference}, which computes output embedding for a set of nodes, and (iii) \textit{sampling inference}, which samples neighborhood during inference (as in training) to trade accuracy for efficiency. DGI also supports out-of-core execution for large graphs. Users can enjoy the efficiency of layer-wise inference and all these functionalities via simple configurations.

We evaluate DGI on 6 real graphs, 5 GNN models with different structures, and various hardware configurations. The results show that DGI consistently outperforms node-wise inference across different settings, and the speedup of DGI over node-wise inference can be up to 1,000x and is usually 10x-100x. Experiments also show that DGI handles different kinds of inference requests efficiently and its inference time scales linearly with the number of layers. Micro experiments show that dynamic batch size control and node reordering are effective in reducing inference time.

%We conducted extensive experiments to evaluate DGI on different graphs, hardware configurations, and GNN models. The results show that DGI is often several orders of magnitude faster than node-wise inference. We also show that DGI enables evaluating complex GNN models on large graphs using hardware with much less CPU memory while current systems easily run out of memory.

%The remainder of the paper is organized as follows. We discuss related works in Section~\ref{sec:related}. Section~\ref{sec:dgi} presents the detail design of \sys. Section~\ref{sec:node-wise and layer-wise} describes layer-wise inference and shows an example of handwritten layer-wise code. Section~\ref{sec:dgi} presents the detail design of \sys. Section~\ref{sec:optimization} shows our optimizations on \sys when the graph datasets are in large-scale. Section~\ref{sec:eval} demonstrates the evaluation of our system. Finally, we present our conclusion in Section~\ref{sec:conclusions}.

%% file: sections/2_layerwise.tex
\section{Layer-wise Inference Approach}\label{sec:node-wise and layer-wise}

\begin{figure}[t]
	\centering
	\includegraphics[width=\linewidth,scale=0.9]{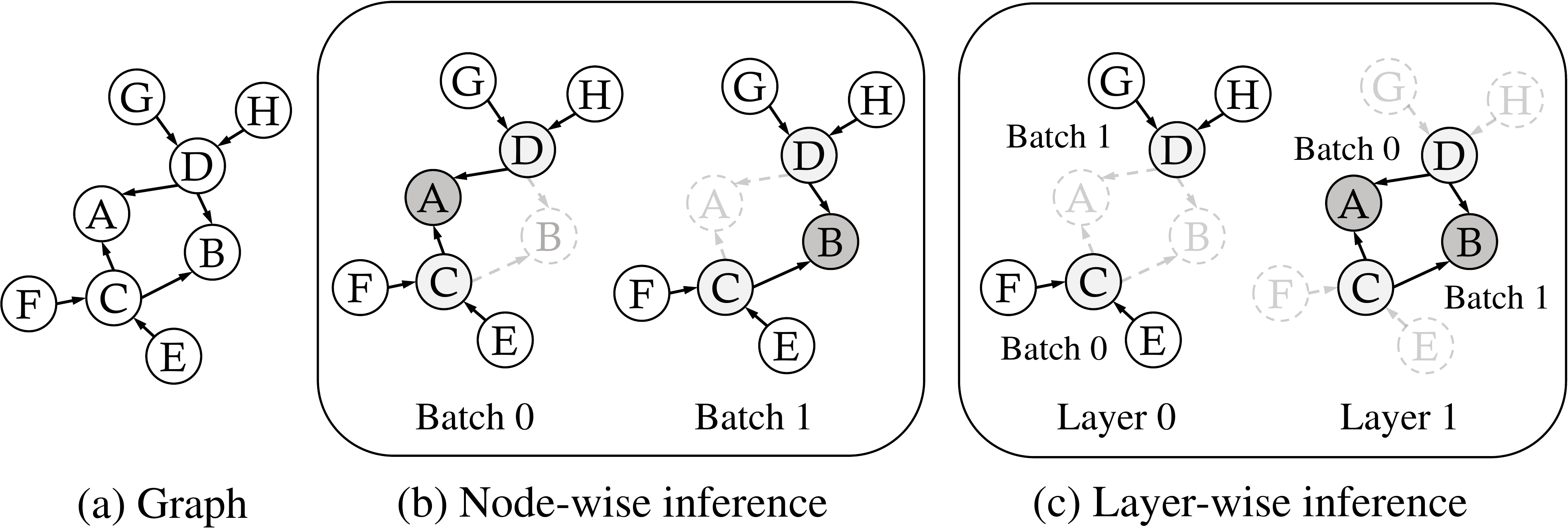}
	\caption{Node-wise and layer-wise inference for a 2-layer GNN model on a toy graph. Both examples illustrate the process to compute output embeddings for node A and B, in two batches with a batch size of 1.}
	\label{fig:infer-scheme}
\end{figure}

Given a graph $G=(V,E)$ and the input (i.e., layer-0) embedding of all nodes $\mathcal{H}^0$, GNN inference computes output embedding using an $L$-layer GNN model. Algorithm~\ref{alg:node-wise} is the pseudo-code of node-wise inference, and Figure~\ref{fig:infer-scheme}(b) provides a running example. In Algorithm~\ref{alg:node-wise}, set $\mathcal{N}^L_v$ contains all $L$-hop neighbors of node $v$, and $\mathcal{H}^0(\mathcal{N}^L_v)$ indicates the input embedding of these neighbors. We assume that input data is stored on an external device (e.g., CPU) and fetched to the computing device (e.g., GPU) for execution in batches. In a batch, node-wise inference takes three steps, i.e., data preparation (Lines~4-6 in Algorithm~\ref{alg:node-wise}), data transfer (Lines~7 and 9), and computation (Line 8). 

% \begin{figure*}[t]
% 	\centering
% 	\includegraphics[width=0.70\textwidth]{figures/illustration.pdf}
% 	\caption{Node-wise and layer-wise inference for a 2-layer GNN model on a toy graph. Both examples illustrate the process to compute output embeddings for node A and B, in two batches with a batch size of 1.}
% 	\label{fig:infer-scheme}
% \end{figure*}

For data preparation, a target node $v$ accesses all its $L$-hop neighbors, which results in a large input set (i.e., neighbor explosion) as the number of neighbors increases quickly with layers. In particular, for super nodes, which have many neighbors and are common in real-world graphs, the input set can constitute a large portion of the entire graph. For this reason, we observe that even with a batch size of 1, node-wise inference can run OOM on GPU for processing large graphs.
%Accessing $L$-hop neighbors is also unfriendly for out-of-core execution when the data is stored on slow memory such as SSD because it requires to read the adjacency list of many nodes and fetch many node embedding. 
Regarding computation, each batch of target nodes is handled independently, and thus different target nodes conduct repetitive computation for their common neighbors. For the example in Figure~\ref{fig:infer-scheme}(b), computing the layer-2 embeddings of both nodes A and B requires the layer-1 embeddings of nodes C and D, and thus the layer-1 embeddings of nodes C and D are computed twice. Empirically, the running time of node-wise inference increases exponentially with the number of layers and constitutes a major part of the training pipeline for multi-layer models.

\begin{algorithm}[H]
    \footnotesize
	\caption{Node-wise Inference for an $L$-layer GNN}
	\label{alg:node-wise}
	\begin{algorithmic}[1]
	    \STATE $\mathcal{H}^L \gets$ empty tensor for holding output embedding
	    \FOR{each batch $V_B$ of nodes in $V$}
	    \STATE $\mathcal{N}\gets$ null neighbor set,  $\mathcal{H}\gets$ null input embedding set
		\FOR{each node $v$ in $V_B$}
		\STATE $\mathcal{N}=\mathcal{N}\cup \mathcal{N}^L_v$, $\mathcal{H}=\mathcal{H}\cup \mathcal{H}^0(\mathcal{N}^L_v)$  
    	\ENDFOR
    	\STATE Copy $\mathcal{N}$ and $\mathcal{H}$ to the computing device
    	\STATE Run Eq.~\eqref{equ:GNN} for each node $v$ in $V_B$ to obtain $\mathcal{H}^L(V_B)$
		\STATE Store $\mathcal{H}^L(V_B)$ to back to $\mathcal{H}^L$ 
		\ENDFOR
	\end{algorithmic}
\end{algorithm}

\begin{algorithm}[H]
    \footnotesize
	\caption{Layer-wise Inference for an $L$-layer GNN}
	\label{alg:layer-wise}
	\begin{algorithmic}[1]
		\FOR{layer $l\in\{1,2,\cdots, L\}$}
		\STATE $\mathcal{H}^{l} \gets$ empty tensor for layer-$l$ embedding
		\FOR{each batch $V_B$ of nodes in $V$}
		\STATE $\mathcal{N}\gets$ null neighbor set, $\mathcal{H}\gets$ null input embedding set
		\FOR {each node $v\in V_B$} 
		\STATE $\mathcal{N}=\mathcal{N}\cup \mathcal{N}_v$, $\mathcal{H}=\mathcal{H}\cup \mathcal{H}^{l-1}(\mathcal{N}_v)$ 
    	\ENDFOR
    	\STATE Copy $\mathcal{N}$ and $\mathcal{H}$ to the computing device
    	\STATE Run layer-$l$ of Eq.~\eqref{equ:GNN} and get the output $\mathcal{H}^l(V_B)$
		\STATE Store $\mathcal{H}^l(V_B)$ (layer-$l$ embedding of batch $V_B$) into $\mathcal{H}^l$
		\ENDFOR
		\STATE Release storage for $\mathcal{H}^{l-1}$ if it is no longer needed 
		\ENDFOR
	\end{algorithmic}
\end{algorithm}

%Conduct inference in layer-wise manner (Figure~\ref{fig:infer-scheme}(c)) can avoid those issues. To infer nodes A and B, it first computes embeddings of nodes C and D in layer 0. The computation is further conducted in two batches. Batch 0 queries one-hops neighbors of node C (node E and F) and extracts their node embeddings, then computes embeddings of node C, and the computation of batch 1 is similar. It then continues to calculate embeddings of nodes A and B using the previously computed embeddings of C and D. Layer-wise inference has several advantages. First, we only need to query one-hops neighbors on the graph, which also result in an acceptable size of neighbor nodes and makes it suitable for out-of-core execution. Second, no embeddings are computed twice during the process. Algorithm~\ref{alg:layer-wise} shows the pseudo-code for layer-wise inference, which the setting is similar to Node-wise. The difference is, it only fetches one-hop neighbor node embeddings each batch, and runs forward of one layer. It repeats for $L$ times and the final layer's output will become the output of the inference.

Algorithm~\ref{alg:layer-wise} is the pseudo-code for layer-wise inference, and Figure~\ref{fig:infer-scheme}(c) shows a running example. In Algorithm~\ref{alg:layer-wise}, $\mathcal{H}^{l}$ contains the layer-$l$ embedding for all nodes, and set $\mathcal{N}_v$ is the 1-hop neighbors of node $v$. Algorithm~\ref{alg:layer-wise} shows that layer-wise inference conducts computation layer by layer, which differs from the node by node scheme of node-wise inference. One advantage of layer-wise inference is small input set as each node only requires its 1-hop neighbors in data preparation (Line 6 of Algorithm~\ref{alg:layer-wise}). The small input set makes layer-wise inference less likely to run OOM for super nodes. Also, small input set allows a large batch size in the inner loop (Line 3 of Algorithm~\ref{alg:layer-wise}), which enables good input sharing. For the example in Figure~\ref{fig:infer-scheme}(c), if the layer-2 embedding of nodes A and B are computed in a batch, then the layer-1 embedding of nodes C and D are loaded to GPU only once and shared by nodes A and B. Another advantage of layer-wise inference is that it completely eliminates repetitive computation. For the example in Figure~\ref{fig:infer-scheme}(c), both nodes A and B require the layer-1 embedding of nodes C and D, which are computed only once in layer-1. 

%This is because the layer-$l$ embedding of the nodes are computed collectively and fetched by layer-$(l+1)$ on demand.              

%Conduct inference in layer-wise manner (Figure~\ref{fig:infer-scheme}(c)) can avoid those issues. To infer nodes A and B, it first computes embeddings of nodes C and D in layer 0. The computation is further conducted in two batches. Batch 0 queries one-hops neighbors of node C (node E and F) and extracts their node embeddings, then computes embeddings of node C, and the computation of batch 1 is similar. It then continues to calculate embeddings of nodes A and B using the previously computed embeddings of C and D. Layer-wise inference has several advantages. First, we only need to query one-hops neighbors on the graph, which also result in an acceptable size of neighbor nodes and makes it suitable for out-of-core execution. Second, no embeddings are computed twice during the process. Algorithm~\ref{alg:layer-wise} shows the pseudo-code for layer-wise inference, which the setting is similar to Node-wise. The difference is, it only fetches one-hop neighbor node embeddings each batch, and runs forward of one layer. It repeats for $L$ times and the final layer's output will become the output of the inference.

We note that layer-wise and node-wise inference are equivalent in semantics because they produce the same output embedding. In Algorithm~\ref{alg:layer-wise}, we present layer-wise inference for full inference, which computes output embedding for all graph nodes without neighbor sampling and is a major use case (e.g., using GNN-generated embedding for applications). Layer-wise inference can also handle other kinds of inference requests. In particular, partial inference computes output embedding for only a subset of the graph nodes and can be used for model evaluation when only some nodes are associated with labels. Here we can conduct BFS of depth $L$ from the target nodes to mark a node set $V^l$, which contains the nodes for which layer-$l$ embeddings need to be computed. Algorithm~\ref{alg:layer-wise} only needs to change the node set from $V$ to $V^l$ in Line 3. Instead of using all neighbors for each node, sampling inference selects some neighbors for aggregation, which is useful when users want to trade accuracy for efficiency. In this case, we mark the node set $V^l$ according to the sampled 1-hop neighbors. 

%Sampling inference can use the layer-wise approach and the difference is that each node only accesses the sampled 1-hop neighbors.
%Sampling inference can use the layer-wise approach similar to partial inference, and the only difference is that each step of BFS only accesses the sampled neighbors.   

\begin{figure*}[t]
	\centering
	\includegraphics[width=0.95\textwidth]{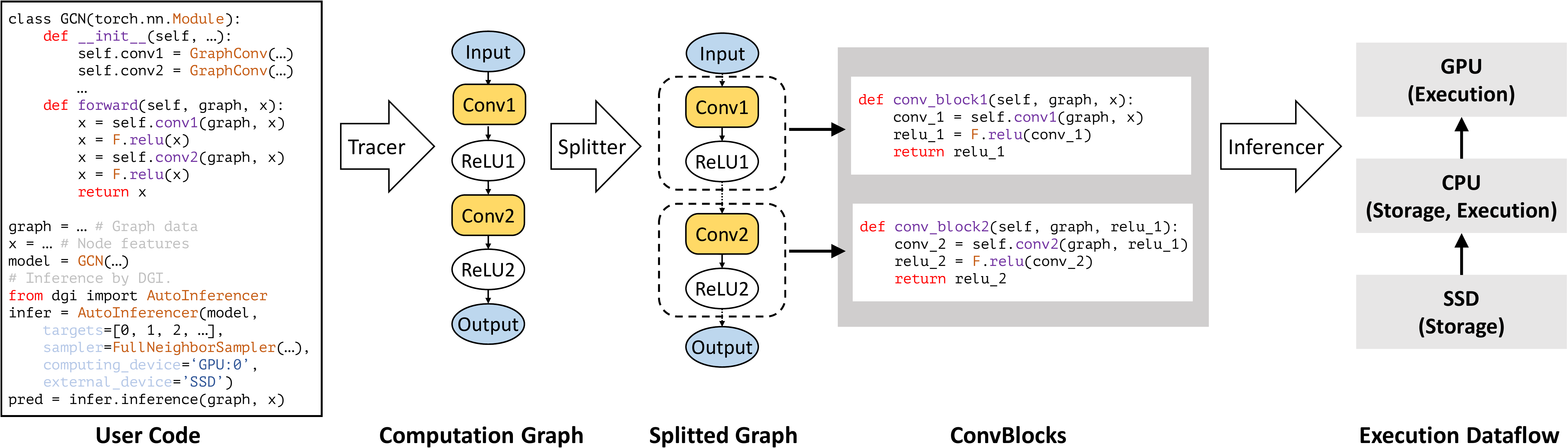}
	\caption{The workflow of \sys.}
	\label{overview}
	\trim
\end{figure*}

Although layer-wise inference has appealing advantages, a drawback is its programming difficulty. Firstly, node-wise inference can directly utilize the training code, while layer-wise inference requires to write separate code for inference. Secondly and more importantly, it can be difficult to decompose a GNN model into layers for inference when it goes beyond a linear stack of layers. For the examples in Figures~\ref{fig:GNN structure}(b) and (c), heterogeneous GNN models have complex connections among the normal operators and graph convolutions, while JKNet has jumping links that connect multiple graph convolution layers. In these cases, a programmer needs to determine a layer partition that is efficient for inference and manage the dependencies of the intermediate embeddings. We show an example in the Appendix for JKNet, whose forward code has only 8 LOCs, but layer-wise inference code needs more than 30 LOCs even after removing some API calls. 
%An expert user of DGL took around 2 hours to get the layer-wise inference code working. 
Additional programming difficulties include: (i) managing the batch size to fit  the memory of the computing device, (ii) handling out-of-core execution for large graphs stored on disk, and (iii) collecting the inference tasks in each layer for partial and sampling inference. These programming difficulties will be more prominent for future GNN models that are more complex~\cite{f2gnn}.

%% file: sections/3_methodology.tex
\section{The DGI System}\label{sec:dgi}

% \begin{lstlisting}[language=Python, caption=Auto inference code example.]
% from dgi import AutoInferencer

% model = GCN(...) # Initialized the GNN model.
% device = torch.device("cuda:0") # Which GPU we use.
% use_uva, use_ssd = True, False # Whether use SSD and UVA.

% # Train the model...
% model.train(graph, *args) # the training function of GNN model.

% # Trace the model forward function.
% infer = AutoInferencer(model, device, use_uva, use_ssd)

% # Takes the same inputs as the forward function.
% pred = infer.inference(graph, *args)
% \end{lstlisting}

%DGI assumes that a single GPU is utilized for inference but extending to multiple GPUs is trivial
%that aggregates the 1-hop neighbors for each node

% Figure~\ref{overview} shows the workflow of

\stitle{API and Overview} \sys is developed to make the use of layer-wise inference easy by automatically translating the training code for execution. We show an example of using \sys in the left part of Figure~\ref{overview}. 
Programmers first import the \textit{AutoInferencer} class from the \textit{dgi} library, and then use the (training code of the) GNN model to initialize an \textit{AutoInferencer} object. 
Parameters to specify include the target nodes that need to compute output embedding (for partial inference, set as all nodes in the graph by default), the neighbor sampling strategy (for sampling inference, use all neighbors by default), the computing device (e.g., a GPU), and where to store the output tensors (e.g., CPU memory and SSD). 
Then the path of the graph data and input node embedding (i.e., graph and x) are fed to the \textit{AutoInferencer} to conduct inference. 
%In the Appendix, we provide a complete example of using DGI to conduct inference for JKNet.
The output embeddings are stored in \textit{pred}. By default, \sys uses a single GPU for inference, and stores data in CPU memory. 

%It is also available to modify the inferencer for other requirements. By default, \sys uses a single GPU for inference, and stores the graph structure and the node feature vectors in CPU memory. 

\sys consists of three key modules, i.e., \textit{tracer}, \textit{splitter}, and \textit{inferencer}, as shown in Figure~\ref{overview}. 
The tracer traces the python code of the GNN model and converts it into a computation graph, which consists of operators such as graph aggregation (denoted as $\mathsf{Conv}$) and activation function.
The splitter uses customized rules to partition the computation graph into \textit{ConvBlocks}, with each ConvBlock corresponding to one layer of the GNN model.
The tracer generates codes for executing each of those ConvBlocks.
% Given the partitioning results of the splitter, the tracer generates python codes for each layer's computation.
The splitter also generates a schema describing the dependencies of the ConvBlocks (i.e., a ConvBlock takes input from which ConvBlocks). 
The inferencer executes the ConvBlocks sequentially (i.e., layer by layer) and runs each ConvBlock in batches. For each batch, the target nodes and their 1-hop neighbors are collected for data preparation, then the batch is transferred to the GPU for computation, and finally the output embeddings of the target nodes are dumped to the CPU. \sys can be configured to fetch data from and store output embedding in SSD when CPU memory is not sufficient.

%Overall the inference process is conducted layer by layer through the GNN model and batch by batch for each layer (i.e., layer-wise inference). 
%For each batch, the target nodes and their one-hop neighbors are collected at first, then the batch is transferred to the GPU for computation, and finally the output embeddings of the target nodes are transferred back to the CPU. 
%\sys also supports storing data in SSD in case the CPU does not have enough memory.

%It is general and can be extended to different scenarios like partial node inference, or using different samplers rather than full neighbors sampler.

\subsection{Tracer}\label{subsec:tracer}

The tracer converts the training code of a GNN model into a computation graph.
We implement DGI tracer using Torch.fx as it generates computation graph in Python format. In particular, Torch.fx uses placeholders called \textit{proxies} to replace the actual tensors and records the operators encountered by the proxies in the forward pass of a model.  
The resultant computation graph is a DAG, where nodes are operators and edges are the dependencies between operators. We classify the operators into two categories, i.e., \textit{normal operator} and \textit{graph convolution operator} ($\mathsf{Conv}$ for short). Normal operators work on the tensor of each individual graph node/edge, and examples include $\mathsf{Add}$, $\mathsf{ReLU}$, and $\mathsf{MatMul}$. A $\mathsf{Conv}$ aggregates the 1-hop neighbors of a graph node to update its embedding. During tracing, DGI tracer recognizes the $\mathsf{Conv}$ operators by matching computation pattern. 
To allow the splitter to determine the execution order of the $\mathsf{Conv}$ operators, we associate an integer $l$ to each $\mathsf{Conv}$ operator to indicate the layer it belongs to. 
In particular, we traverse the computation graph from the input, and mark the $\mathsf{Conv}$ operators that have no preceding $\mathsf{Conv}$ operators as layer 1. For a $\mathsf{Conv}$ operator with preceding $\mathsf{Conv}$ operators (called precursors), its layer is marked as one plus the maximum layer of its precursors.

\subsection{Splitter}\label{subsec:splitter}

\begin{figure}[t]
	\centering
\includegraphics[width=\linewidth,scale=0.95]{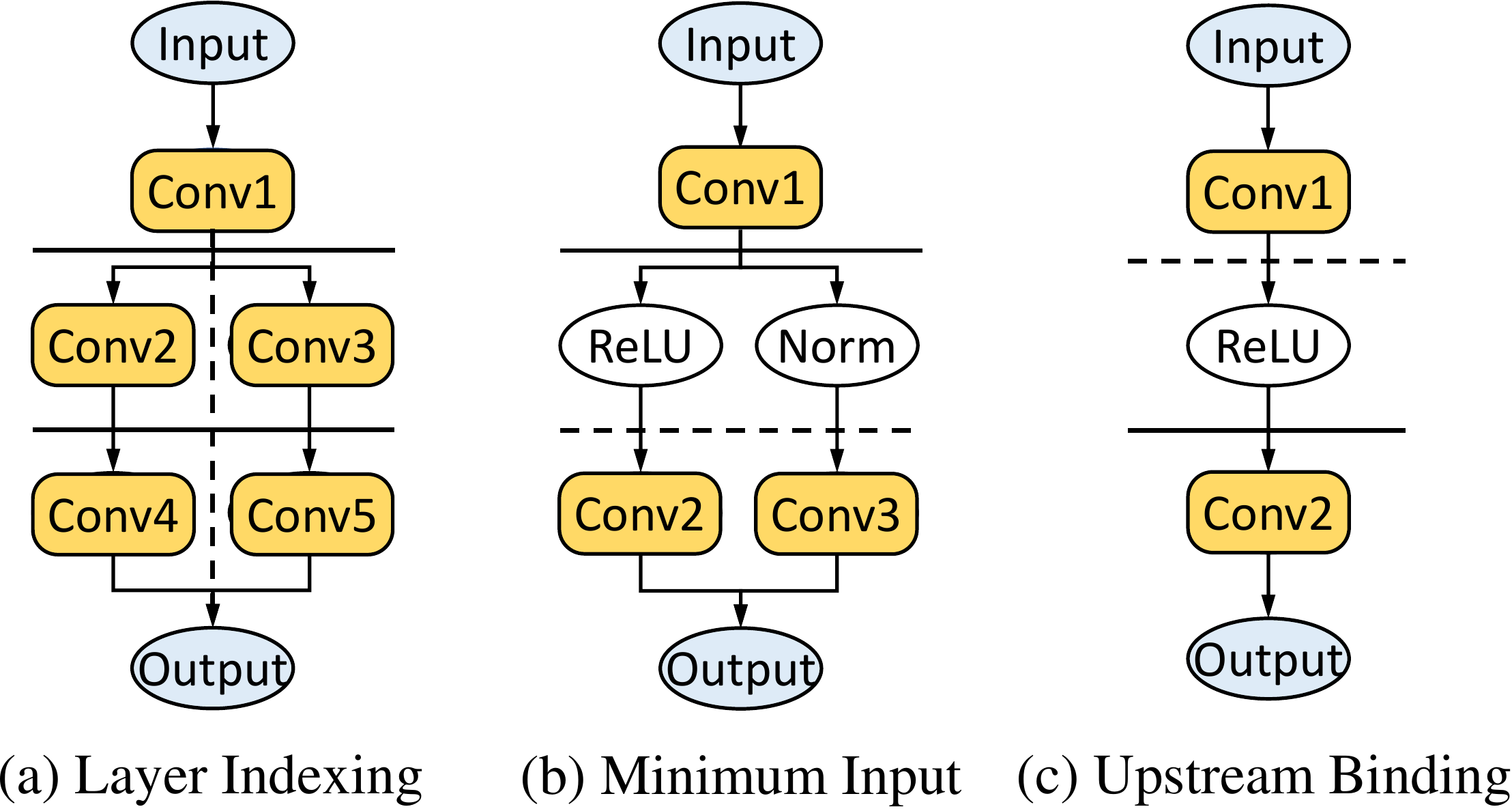}
	\caption{Illustrations of the rules for computation graph partitioning. The solid line denotes the partitioning adopted by \sys while the dotted line is an inferior alternative.}
	\label{fig:split}
	\trim
\end{figure}

% In addition to $\mathsf{Conv}$ operators, a GNN model may contain many other operators such as $\mathsf{ReLU}$ and $\mathsf{MatMul}$. 
% Thus, the splitter needs to decide how to pack these operators with the $\mathsf{Conv}$ operators to form ConvBlocks, each ConvBlock corresponding to a layer in the GNN model. 
% In particular, the splitter partitions the computation graph of the GNN model into sub-graphs and each sub-graph is treated as a ConvBlock. 
% The following three rules are adopted for computation graph partitioning.

%To conduct efficient inference for large graphs on limited GPU memory, 
The \sys splitter partitions the computation graph of a GNN model into sub-graphs, and each sub-graph is treated as a ConvBlock (i.e., layer). 
%When computing the ConvBlock of a batch of nodes, we move its neighbor nodes embedding from CPU to GPU and transfer the output node embedding back to CPU.
We adopt the following three rules for computation graph partitioning.
%to facilitate efficient layer-wise execution. 

% to avoid redundant computation and minimize data transfer between CPU and GPU.

%The rational is to partition two consecutive $\mathsf{Conv}$ operator
\stitle{Layer indexing}
It means that $\mathsf{Conv}$ operators are organized into the same ConvBlock if and only if they have the same layer counter $l$. One such example is shown in Figure~\ref{fig:split}(a), where $\mathsf{Conv2}$ and $\mathsf{Conv3}$ are in the same ConvBlock because their layer counters are both 2. 
As the $\mathsf{Conv}$ operators in a ConvBlock have the same layer counter, the layer indexing rule ensures that a target node only needs its 1-hop neighbor nodes to execute each ConvBlock. 
An alternative is to treat each $\mathsf{Conv}$ operator as a ConvBlock (which also satisfies layer-wise inference), for example, $\mathsf{Conv2}$ and $\mathsf{Conv3}$ could be two ConvBlocks. 
However, this partitioning misses possible opportunities for \textit{input sharing} among ConvBlocks. 
For the example in Figure~\ref{fig:split}(a), both $\mathsf{Conv2}$ and $\mathsf{Conv3}$ take the output embedding of $\mathsf{Conv1}$ as input, and thus can share input embeddings when they are loaded to the GPU. 
When treated as two separate ConvBlocks, $\mathsf{Conv2}$ and $\mathsf{Conv3}$ need to load the embedding generated by $\mathsf{Conv1}$ individually. 

%This rule minimizes the data transfer for the node embedding between CPU and GPU.
%as neighbor nodes
\stitle{Minimum input}
It means that when there are multiple ways to partition two consecutive ConvBlocks, choose the way that yields the minimum number of input tensors for the downstream ConvBlock.
Note that when an edge in the computation graph is selected as a cut, the output node embedding of the upstream ConvBlock is saved on the CPU and loaded to the GPU for the downstream ConvBlock. 
For the example in Figure~\ref{fig:split}(b), both the dotted line and the solid line satisfy the layer indexing rule.
We choose the solid line as it only requires each neighbor node to transfer one embedding vector to the GPU for the second ConvBlock (by applying the $\mathsf{ReLU}$ and $\mathsf{Norm}$ operators to a node embedding after it is loaded to the GPU). 
In contrast, if the dotted line is used, each neighbor node needs to transfer two embedding vectors to the GPU (as $\mathsf{ReLU}$ and $\mathsf{Norm}$ are applied to the output of the upstream ConvBlock). 
Thus, the minimum input rule minimizes the neighbor node data required by each ConvBlock for small data transfer cost.

%, and thus achieves a small cost for moving data to the GPU. 

\stitle{Upstream binding}
It means that a normal operator should be grouped with the upstream ConvBlock when there are multiple partitioning schemes to minimize data transfer. 
For example, in Figure~\ref{fig:split}(c), we choose the solid line as the cut as it puts the $\mathsf{ReLU}$ operator in the first ConvBlock. 
This is because if we group normal operators with the downstream ConvBlock (i.e., using the dotted line), there will be repetitive computation. 
The reason is that a node usually serves as a neighbor node for multiple target nodes (e.g., node C is used by nodes A and B in Figure~\ref{fig:infer-scheme}), and if the dotted line is used, each target node needs to apply the normal operator on its neighbor nodes individually. 
In contrast, the solid line applies the normal operator only once for each neighbor node when generating output for the upstream ConvBlock. Thus, the upstream binding rule avoids repetitive computation.

%This rule indicates that we prefer executing normal operators on the output tensor rather than the input tensor for a ConvBlock to avoid repetitive computation. 

Based on these rules, we adopt a 3-step procedure to partition the computation graph into ConvBlocks. 
First, we determine the ConvBlock of each $\mathsf{Conv}$ operator according to the layer counters of the $\mathsf{Conv}$ operators (i.e., layer indexing). 
Second, starting from the $\mathsf{Conv}$ operators of the same ConvBlock, we traverse their preceding operators in the computation graph to find the partition points with the minimum number of inputs (i.e., minimum input). 
If there are multiple minimum input points, we choose the one that groups the largest number of normal operators into the upstream ConvBlocks (i.e., upstream binding). 
%Finally, we group the other normal operators into the relatively upstream ConvBlocks.
Note that minimum input and upstream binding conflict sometimes (e.g., Figure~\ref{fig:split}(b)). 
We prioritize minimum input as repetitive data loading is more expensive than the repetitive execution of normal operators.   

Given the partitioning results of the splitter, the tracer registers each ConvBlock to a python function for execution. 
As a ConvBlock may take input from multiple ConvBlocks (e.g., JKNet), we use a schema to record the dependencies among the ConvBlocks, and each ConvBlock fetches input embedding according to the schema. 
We store the output embedding of each ConvBlock in a separate map, and the output of a ConvBlock is deleted when there are no ConvBlocks depending on it.

%are depending on it.  

\subsection{Inferencer}\label{subsec:inferencer}

The inferencer executes the ConvBlocks sequentially and runs each ConvBlock batch by batch. To execute a ConvBlock, the inferencer first allocates memory to hold its output node embeddings. For each batch, execution takes four steps: (i) \textit{data preparation}, a set of target nodes $V_B$ are selected for the batch, and for each node $v\in V_B$, its 1-hop in-neighbors are collected into the input set along with their embeddings (we modify DGL dataloader for this purpose); (ii) \textit{data transfer}, input set of batch $V_B$ is transferred to the GPU; (iii) \textit{computation}, the output embeddings for nodes in $V_B$ are computed on the GPU; and (iv) \textit{embedding dumping}, the output embeddings are transferred back to CPU. To achieve a short inference time, the inferencer should execute each ConvBlock in a small number of batches while avoiding OOM, and we discuss node batching in Section~\ref{sec:optimization}. 

%\begin{lstlisting}[language=Python, caption=Compute function for the basic inferencer., label=code:compute]
%def compute(self, conv_block, *args):
%    # Create empty tensors for outputs.
%    rets = create_empty_tensors(conv_block)
%    # Initialize the dataloader.
%    dataloader = NodeDataLoader(
%        self.input_graph,
%        self.nids,
%        MultiLayerFullNeighborSampler(1),
%        batch_size=self.batch_size,
%        device=self.device)
%
%    for in_nodes, out_nodes, blocks in dataloader:
%        # Get input node tensors, and transfer data 
%        # to target device.
%        new_args = get_new_arg_input(
%            args, in_nodes, blocks[0], self.device)
%        # Compute the ConvBlock.
%        out_vals = conv_block(*new_args)
%        # Update result to rets.
%        rets = update_ret_output(
%            out_vals, rets, out_nodes, blocks)
%
%    return rets
%\end{lstlisting}

%From the users' perspective, the \textit{compute} function takes a ConvBlock and its relative arguments as inputs, and returns the output tensors. To support different scenarios or advantage usage, users only need to extend the inferencer class and overwrite the \textit{compute} function. For example, users can replace the full neighbor sampler with a fan-out neighbor sampler to support GNN inference by sampling neighbors. Users can also add steps to arrange tensors in pinned memory to accelerate data transfer time. For the next section, we will introduce several optimizations we adopted in \sys.

For full inference, the target nodes of all ConvBlocks are the complete graph node set. For partial inference, the inferencer annotates the target nodes for each ConvBlock before execution. In particular, for an $L$-layer GNN model, the $L\textsuperscript{th}$ layer target node set $V^L$ contains the target nodes specified by the user. The $(L-1)\textsuperscript{th}$ layer target node set $V^{L-1}$ contains the 1-hop in-neighbors of nodes in $V^L$. The annotation process goes on recursively until obtaining $V^1$, which is the target node set for layer-1. For a ConvBlock on layer-$l$ ($l\in[1, L]$), the inferencer only computes output embedding for nodes in $V^l$. Annotation is similar for sampling inference, where we annotate the sampled neighbors for nodes in $V^{l+1}$ as $V^{l}$. We observe that the annotation process for $V^{l}$ is expensive and the resultant target node sets contain most of the graph nodes if $V^{l+1}$ ($l<L$) is already very large. As an optimization, we skip the annotation and directly assign all nodes to $V^l$ when $|V^{l+1}|\times d_{avg}\ge n$, where $d_{avg}$ is the average  degree and $n$ is the number of graph nodes.

%% file: sections/4_optimizations.tex
\section{Handle Large Graphs}
\label{sec:optimization}

Real-world graphs can be large and cause two problems for inference. First, for a layer, target nodes and input embedding may not fit in GPU memory, which requires to organize the target nodes into batches for computation and control the batch size. Second, graph data and intermediate embedding may not fit in CPU memory, which requires disk storage. We discuss related optimizations in this part.   

%Real-world graph datasets are usually large-scale. We further proposed several optimizations in \sys for dealing with those large graphs. To accelerate inference, we use dynamic batch size control and node reorder to increase node sharing among target nodes. We further use SSD for storage to avoid the problem caused by the limited CPU memory.

\subsection{Dynamic Batch Size Control}\label{subsec:dynamic batch size}

Batch size refers to the number of target nodes in a batch, and a good batch size should avoid exceeding GPU memory and be as large as possible. This is because a large batch size allows better input sharing (i.e., multiple target nodes share the same input node embedding) and thus reduces data transfer. However, determining a good batch size is challenging. First, it is difficult to estimate the memory consumption of neural network models due to reasons such as temporary tensors~\citep{dnnmem}, and thus even experienced programmers can only choose a safe but small batch size before execution. Second, a static batch size may not be suitable for all batches. For example, if a batch mainly contains nodes with large in-degrees, a small batch size is needed when each target node has many neighbors while a large batch size should be used if a batch mainly contains nodes with small in-degrees. 
%Batch size configuration is even more complicated with input neighbor node sharing among the target nodes.

Considering the challenges, we use a dynamic strategy to configure the batch size according to run-time statistics. In particular, we observe that the number of target nodes (i.e., batch size, denoted as $n_t$) and the total in-degrees of the target nodes (denoted as $n_i$) are closely related to the GPU memory consumption of a batch, and thus keep adding target nodes to a batch until either the node count threshold or edge count threshold is exceeded. The two thresholds are initialized as safe values to avoid OOM and adjusted in the inference process. Specifically, for the $k\textsuperscript{th}$ batch, we measure its peak memory consumption as $M_k$, and adjust the thresholds for the next batch as
\begin{equation}
	r=M/M_k, \quad n_t^{k+1}=r\times n_t^k, \quad n_i^{k+1}=r\times n_i^k,
\end{equation} 
where $n_t^k$ (resp. $n_i^k$) is the node count (resp. edge count) threshold for the $k\textsuperscript{th}$ batch. $M$ is the expected memory consumption and set as 90\% of the available GPU memory. The rationale is to increase batch size if memory is underutilized. If a batch runs OOM, we reduce the thresholds to half of their previous values, and \sys re-conducts the batch. For quick node batching, we precompute prefix sum for the in-degrees of the nodes such that binary search can be used to find a batch of target nodes meeting the thresholds. 
%Using the actual number of in-neighbor nodes of a batch allows more accurate control of batch size. However, duplicating the neighbor nodes when adding each target node is expensive, and experiment results show that using node and in-degree count yields good performance. 

\begin{table*}[t]
	\footnotesize
	\caption{Statistics of the graphs used in the experiments.}
	\label{tab:evaluation-datasets}
	\centering
	\fontsize{8}{10}\selectfont 
	\begin{tabular}{cccccc}
		\toprule
		\textbf{Name} & \textbf{Nodes} & \textbf{Edges} & \textbf{Input Dim.} & \textbf{Average Degree} & \textbf{Total Size} \\
		\midrule
		Livejournal1 & 4,847,571 & 86,220,856 & 100 & 17.8 & 2.45 GB
		\\
		OGBN-Products & 2,449,029  & 126,167,309 & 100 & 51.5 & 2.11 GB    \\
		OGBN-Papers100M & 111,059,956 & 1,726,745,828 & 100 & 15.5 & 64.96 GB \\
		Friendster & 65,608,366	& 3,612,134,270 & 128 & 55.0	& 58.20 GB     \\
		\midrule
		OGBN-MAG (Hetero) & 1,939,743 & 21,111,007 & 128 & 10.9 & 0.94 GB
		\\
		OGBN-MAG240M (Hetero) & 244,160,499 & 3,456,728,464 & 768 & 14.2 & 200.39 GB
		\\
		\bottomrule
	\end{tabular}
\trim
\end{table*}

\begin{table*}[t]
	\centering
	% 	\fontsize{8}{11}\selectfont    %{字体尺寸}{行距}
	\footnotesize
	\caption{Execution time of node-wise inference and \sys (in seconds). For large datasets (\textit{OGBN-Papers100M}, \textit{Friendster} and \textit{OGBN-MAG240M}), node-wise inference may encounter a super large input node set for 3-hop neighbors, which results in OOM. So here we use 2-layer GNN models for those large datasets. When node-wise inference runs for more than 10,000s, we estimate its execution time by $t\times\frac{1}{\alpha}$, where $t$ is the elapsed execution time and $\alpha$ is the percentage of processed target nodes.}
	\label{tab:main comparison}
	\fontsize{8}{9}\selectfont 
	\begin{tabular}{c|ccc|ccc|cc}
		%\toprule
		%  \cr
		\cmidrule(lr){1-9}
		&
		\multicolumn{3}{c}{\textbf{OGBN-Products}}&\multicolumn{3}{c}{\textbf{Livejournal1}}&\multicolumn{2}{c}{\textbf{OGBN-MAG}} \cr
		\cmidrule(lr){1-9}
		GNN Models (3 layers) & GCN & GAT & JKNet & GCN & GAT & JKNet & RGCN & HGT \cr
		\cmidrule(lr){1-9}
		Node-wise (V100-16GB) & 7110 & 9740 & 7150 & 4330 & 5220 & 4890 & 208 & 263 \cr
		\sys(V100-16GB) & 6.08 & 8.96 & 6.70 & 11.0 & 16.9 & 13.5 & 19.7 & 22.0 \cr
		Node-wise (V100-32GB) & 1690 & 1970 & 1760 & 1630 & 2110 & 1990 & 143 & 204 \cr
		\sys(V100-32GB) & 4.51 & 6.09 & 4.20 & 6.32 & 10.5 & 7.10 & 12.8 & 14.9 \cr
		\bottomrule
	\end{tabular}\vspace{0cm}
	\begin{tabular}{c|ccc|ccc|cc}
		%\toprule
		%  \cr
		\cmidrule(lr){1-9}
		&
		\multicolumn{3}{c}{\textbf{OGBN-Papers100M}}&\multicolumn{3}{c}{\textbf{Friendster}}&\multicolumn{2}{c}{\textbf{OGBN-MAG240M}} \cr
		\cmidrule(lr){1-9}
		GNN Models (2 layers) & GCN & GAT & JKNet & GCN & GAT & JKNet & RGCN & HGT \cr
		\cmidrule(lr){1-9}
		Node-wise (V100-16GB) & 3620 & 3980 & 3470 & 57300 & 81900 & 63700 & 29400 & 143000 \cr
		\sys(V100-16GB) & 370 & 607 & 343 & 425 & 741 & 435 & 999 & 1580 \cr
		Node-wise (V100-32GB) & 1950 & 3070 & 1940 & 48700 & 64900 & 33600 & 18200 & 70200 \cr
		\sys(V100-32GB) & 177 & 397 & 175 & 238 & 465 & 231 & 643 & 1010 \cr
		\bottomrule
	\end{tabular}\vspace{0cm}
	\trim
\end{table*}

%that may be significantly longer than the GNN inference time
\subsection{Node Reorder}\label{subsec:node reorder}
A naive solution to node batching is to randomly group the target nodes into batches. This is sub-optimal as some target nodes can share input embedding to reduce data transfer. Take the graph in Figure~\ref{fig:infer-scheme} for example and consider a batch size of 2. If nodes A and C form a batch, 4 neighbor nodes are required (i.e., node A requires C and D, node C requires E and F); however, if nodes A and B form a batch, only 2 neighbor nodes are required as the in-neighbors of both A and B are C and D. Therefore, we should organize target nodes sharing common neighbors into the same batch. For this purpose, the natural solution is to partition the graph into strongly connected components and treat nodes in each component as a batch. However, graph partitioning tools such as METIS~\citep{metis} and KaFFPaE~\citep{KaFFPaE} have long running time. Node reordering algorithms like Rabbit~\citep{rabbit} and Gorder~\citep{gorder} run fast but are designed for improving cache hit and perform poorly for our purpose (see Section~\ref{subsec:exp micro}). Thus, we use a lightweight strategy to renumber the nodes and take nodes with consecutive ids as a batch. In particular, RCMK~\citep{rcmk}, a BFS-based algorithm, is used to enumerate the nodes, and ids are assigned to the nodes in their enumeration order. The rationale is that nodes adjacent in the BFS order are likely to share common neighbors.

\subsection{SSD Support}\label{subsec:ssd}
%When the graph data is large or a cheap machine is utilized, the graph data or output embedding may not fit in CPU memory. 
\sys allows to store graph data and intermediate embeddings in SSD and automatically loads them to conduct inference. In particular, we use numpy~\citep{numpy} \textit{memmap} to map SSD as part of CPU memory. A new graph type called \textit{SSDGraph} is utilized to store the graph structure in CSC format on SSD, and an \textit{SSDNeighborSampler} is developed to extract the adjacency lists of the target nodes from \textit{SSDGraph} to CPU memory for each batch. The required input embeddings are also loaded to CPU memory using \textit{memmap}. After the GPU finishes a batch, \sys writes the output embeddings back to SSD.

%The batch is finally transferred to the GPU for inference, and the results are written back to the SSD. 

%% file: sections/5_evaluation.tex
\section{Experimental Evaluation} \label{sec:eval}

We introduce the experiment settings in Section~\ref{subsec:exp setting}, compare \sys with node-wise inference in Section~\ref{subsec:exp main}, and evaluate the optimizations of \sys in Section~\ref{subsec:exp micro}.

\input{sections/6_subsections/1_settings.tex}

\input{sections/6_subsections/2_nodewise.tex}

\input{sections/6_subsections/3_optimizations.tex}

%% file: sections/6_subsections/1_settings.tex
\subsection{Experiment Settings}\label{subsec:exp setting}

%\stitle{Models and datasets} 

We use 3 popular homogeneous GNN models, i.e., GCN~\citep{GCN}, GAT~\citep{GAT}, and JKNet~\citep{JKNet}, and 2 heterogeneous GNN models, i.e., RGCN~\citep{rgcn} and HGT~\citep{hgt}. GCN uses mean pooling to aggregate neighbor embeddings while GAT adopts multi-head attention (we use 2 attention heads). The final layer of JKNet aggregates neighbor embeddings from all preceding layers. 
%RGCN is based on GCN and considers node labels in aggregation. HGT uses extra parameters to introduce heterogeneous attention over each type of edge. 
Both RGCN and HGT work on heterogeneous graphs and aggregate neighbor embeddings based on certain \textit{src-edge-dst} type triplets. 
%RGCN uses mean aggregation while HGT uses attention-based aggregation. 
For all models, the dimension of intermediate embedding is set as 128. We conduct experiments on 6 popular and public graphs, i.e., \textit{OGBN-Products}, \textit{OGBN-MAG}, \textit{OGBN-Papers100M}, \textit{OGBN-MAG240M} from the Open Graph Benchmark~\citep{ogbn}, and \textit{Friendster}, \textit{Livejournal1} from the Stanford Network Analysis Platform~\citep{snap}. Table~\ref{tab:evaluation-datasets} reports the statistics of the graphs. Among them, \textit{OGBN-MAG} and \textit{OGBN-MAG240M} are heterogeneous graphs used to experiment heterogeneous GNNs.

%\stitle{Hardware} 

Both \sys and the baselines are implemented on top of DGL. We use two different machines for the experiments. One machine has an Intel(R) Xeon(R) E5-2686 CPU with 96 cores, 488 GB main memory, and one NVIDIA V100 GPU with 16GB HBM (denoted as V100-16GB). The other machine has an Intel(R) Xeon(R) Gold 6126 CPU with 24 cores, 1.5TB main memory, and one NVIDIA V100 with 32GB GPU HBM (denoted as V100-32GB). The two machines have sufficient main memory to store data and intermediate embedding for all graphs, and we use them to explore the influence of GPU memory. We use the execution time of inference as the main performance metric and keep 3 effective numbers in the results. For partial inference, we randomly sample some nodes from the graph as the target nodes. For sampling inference, we use a fan-out of 10 for all layers following the GraphSage~\citep{SAGE} implementation in DGL~\citep{dgl}, which means that each target node samples 10 neighbors. To run node-wise inference, we use a grid search to find the batch size that minimizes inference time.

%% file: sections/6_subsections/2_nodewise.tex
\begin{table*}[h]
	\centering
		%\fontsize{7}{9}\selectfont    %{字体尺寸}{行距}
	\footnotesize
	\caption{Execution time of node-wise inference and \sys (in seconds) for full inference the changing the number of layers. The graph is \textit{OGBN-Products}, ``$\star$" means that JKNet requires at least 2 layers, and "OOM" indicates that an out-of-memory exception was raised.}
	\label{tab:layer-wise comparison}
	\fontsize{8}{9}\selectfont 
	\begin{tabular}{c|cccc|cccc|cccc}
		%\toprule
		%  \cr
		\cmidrule(lr){1-13}
		&
		\multicolumn{4}{c}{\textbf{GCN}}&\multicolumn{4}{c}{\textbf{GAT}}&\multicolumn{4}{c}{\textbf{JKNet}} \cr
		\cmidrule(lr){1-13}
		\# of Layers & 1 & 2 & 3 & 4 & 1 & 2 & 3 & 4 & 1 & 2 & 3 & 4 \cr
		\cmidrule(lr){1-13}
		Node-wise (V100-16GB) & 3.16 & 153 & 7110 & OOM & 3.03 & 189 & 8740 & OOM & $\star$ & 152 & 7150 & OOM \cr
		\sys(V100-16GB) & 2.60 & 4.37 & 6.10 & 9.79 & 2.62 & 5.68 & 8.96 & 14.5 & $\star$ & 4.31 & 6.70 & 11.9 \cr
		Node-wise (V100-32GB) & 2.56 & 26.8 & 1690 & 69400 & 2.63 & 29.4 & 1970 & 122000 & $\star$ & 70.2 & 1760 & 90900 \cr
		\sys(V100-32GB) & 1.75 & 3.32 & 4.51 & 5.40 & 1.95 & 4.76 & 6.09 & 8.10 & $\star$ & 3.29 & 4.20 & 5.71 \cr
		\bottomrule
	\end{tabular}\vspace{0cm}
\trim
\end{table*}

\subsection{Main Results}\label{subsec:exp main}

\stitle{Full inference} We report the execution time of node-wise inference and \sys for full inference in Table~\ref{tab:main comparison}, which lead to several observations. Firstly, \sys consistently outperforms node-wise inference for all datasets, models and hardware configurations, and the speedup ranges from 10.6x to over 1000x. Considering the homogeneous GNNs (i.e., GCN, GAT and JKNet), \sys yields larger speedup for 3-layer models and dense graphs (i.e., having large average degrees). In particular, 3-layer models all observe over 2 orders of magnitude speedup while 2-layer models may have 10x speedups. For 2-layer models, the speedup of \sys is 10x for \textit{OGBN-Papers100M} (with an average degree of 15.5) and 100x for \textit{Friendster} (with an average degree of 55). This is because more layers and denser graphs make node-wise inference access more neighbor nodes for each target node, which aggravates the neighbor explosion problem. Secondly, both \sys and node-wise inference run faster on V100-32GB than V100-16GB because larger GPU memory allows larger batch size, and thus there are more opportunities for input sharing. Thirdly, \sys  generally observes smaller speedups for the heterogeneous GNNs (i.e., RGCN and HGT) than the homogeneous GNNs. This is because each Conv only aggregates neighbors of a specific type in heterogeneous GNNs, and thus the neighbor explosion problem of node-wise inference is alleviated. Comparing the results on \textit{OGBN-MAG} and \textit{OGBN-MAG240M}, we also observe that \sys has larger speedup on bigger and denser graphs for heterogeneous GNNs.

%We use 3-layer models for the 3 small graphs and 2-layer models for the 3 large graphs as node-wise inference goes OOM for 3-layer models even with a batch size of 1. This is because graphs usually contain super-nodes with many neighbors, and the 3-hop neighbors of these super-nodes are numerous for large graphs. Our \sys is free from this problem as it only accesses the 1-hop neighbors for each node. To complete the experiments in reasonable time, we estimate the running time of node-wise inference when it runs for more than 10,000 seconds. In particular, denote $t$ as the elapsed running time and $\alpha$ as the percentage of processed target nodes, we estimate the total execution time as $t/\alpha$.             

%\textit{OGBN-Products}, \textit{Livejournal1} and \textit{OGBN-MAG} use three layers GNN models. For large datasets (\textit{OGBN-Papers100M}, \textit{Friendster} and \textit{OGBN-MAG240M}), the node wise approach may contains a super large input nodes set for three hops neighbors. So here we use two layers GNN models for those large datasets. When node-wise inference runs for more than 10,000s, we estimate its execution time by $t\times\frac{1}{\alpha}$, where $t$ is the elapsed execution time and $\alpha$ is the percentage of processed target nodes.

\begin{figure}[t]
	\centering
	\includegraphics[width=\linewidth,scale=0.9]{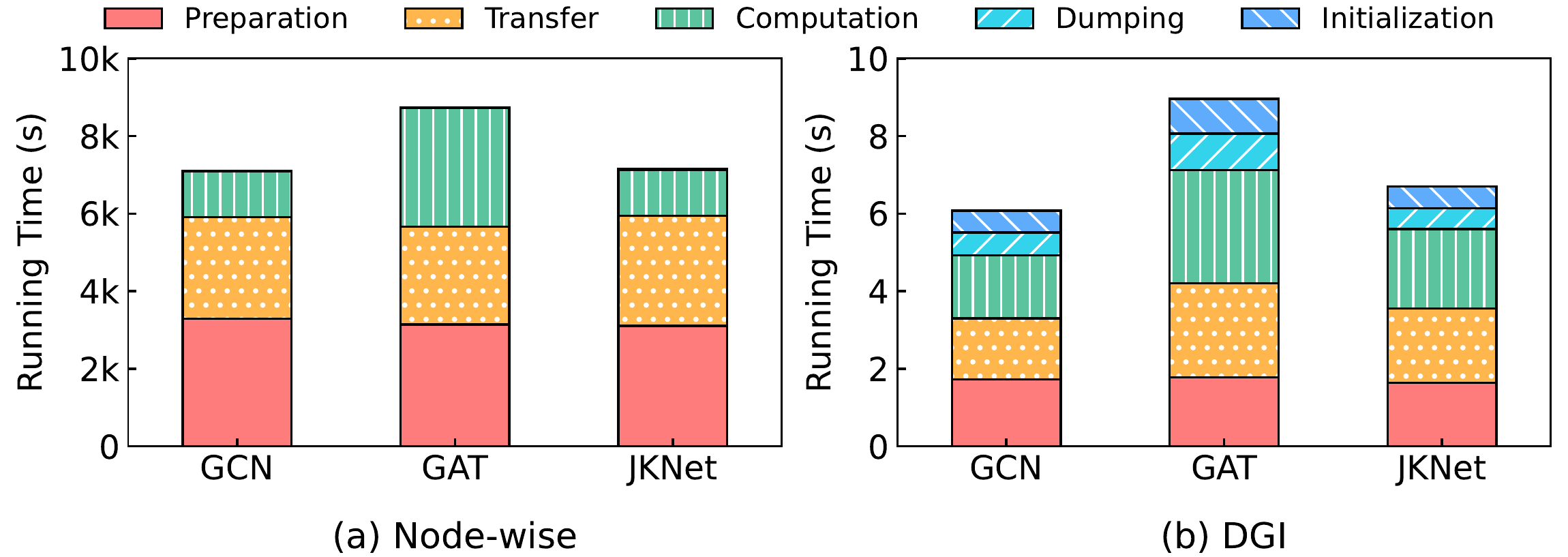}
	\caption{Decomposing the running time of node-wise inference and \sys for full inference. The results are obtained for the \textit{OGBN-Products} graph on V100-16GB with 3-layer models.}
	\label{fig:decompose}
	\trim
\end{figure}

\begin{figure*}
	\centering
	\includegraphics[width=\textwidth]{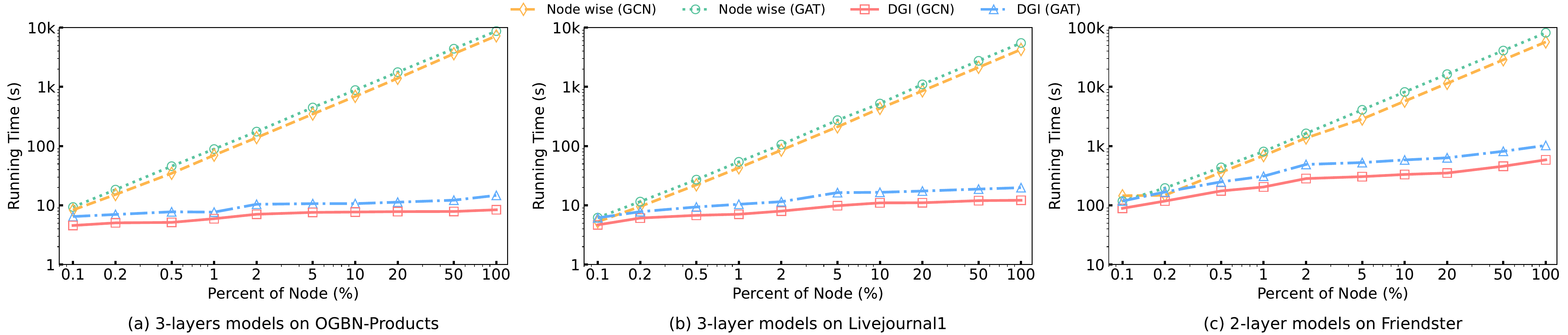}
	\caption{Running time of \sys and node-wise for partial inference (on V100-16GB GPU). Note that both axes use log scale.}
	\label{fig:partial}
	\trim
\end{figure*}

In Table~\ref{tab:layer-wise comparison}, we compare node-wise inference and \sys by changing the number of layers in the GNN models. The results show that the running time of node-wise inference increases quickly with the number of layers. This is because node-wise inference accesses $L$-hop neighbors for $L$-layer models, and the number of neighbors increases quickly with $L$. In contrast, the execution time of \sys increases almost linearly with the number of layers as layer-wise inference only accesses one-hop neighbors and is free from the neighbor explosion problem. For 1-layer models,  node-wise inference and the layer-wise inference of \sys are equivalent, and thus they yield comparable running time. For 4-layer models, \sys usually outperforms node-wise inference by more than 1000x, and node-wise inference easily goes OOM.

% even for the small \textit{OGBN-Products} graph. 

%, which includes \textit{data preparation}, \textit{data transfer}, \textit{computation}, \textit{embedding dumping}, and \textit{initialization}. We have introduced the first 4 parts in Section~\ref{subsec:inferencer}, and initialization means initializing the dataloader and creating empty tensors for output embeddings.
To understand the speedup of \sys over node-wise inference, we decompose their running time in Figure~\ref{fig:decompose}. The first 4 time comsuptions are introduced in Section~\ref{subsec:inferencer}, and \textit{initialization} means initializing dataloader and creating empty tensors for output embeddings. Notice that the y-axis uses different scales for \sys and node-wise inference. The results show that node-wise inference uses significantly longer time than \sys for data preparation, data transfer and GNN forward, which make the time for embedding dumping and initialization negligible. In particular, each of the three parts (i.e., preparation, data transfer and GNN forward) takes thousands of seconds for node-wise inference but only several seconds for \sys. This is because node-wise inference accesses $L$-hops neighbors for data preparation, and the large input set results in a small batch size, which makes input and computation sharing difficult. In contrast, data preparation is quick for \sys as it only accesses 1-hop neighbors, and small input set enables large batch size for good input sharing. 
%For example, when running GCN on the \textit{Friendster} graph, node-wise can only access 2-hops neighbors with a batch size of 400 while \sys (without dynamic batch size control) can use a batch size of 20000.

%We can also figure out that both in \sys and node-wise, the multi-heads mechanism in GAT brings more computation in GNN forward compare to GCN. In \sys, JKNet and GAT both have longer data transfer time than GCN -- JKNet requires one aggregation layer for all hidden embeddings and GAT with multi-heads have a larger dimension for hidden embeddings.

\stitle{Partial inference} We compare node-wise inference and \sys for partial inference in Figure~\ref{fig:partial}. The results show that when increasing the number of target nodes, the inference time increases linearly for node-wise inference but sub-linearly for \sys. This is because \sys effectively shares common computation and input nodes for different target nodes while sharing is difficult for node-wise inference due to the small batch size caused by neighbor explosion. While \sys and node-wise inference perform comparably when the target node set is small (e.g., 0.1\% of all graph nodes), the speedup of \sys over node-wise inference increases with the number of target nodes. This is because more target nodes enables \sys to better share input and computation.

%While \sys only perform low runtime all the time. When the target node sets are small (like 0.1\% of nodes), the computation time is also small, and the performance of \sys and node-wise are similar since they all need to do the initialization step. With the increase of target node sets, the overhead of \sys only increases slightly because layer-wise avoid repeated calculation, which does not generate large read sets as node-wise facing.

\stitle{Sampling inference} Instead of using exact inference with the complete neighbor set, some users may use neighbour sampling for inference to trade accuracy for efficiency. We compare \sys and node-wise inference for sampling inference in Table~\ref{tab:sampled}. As sampling mitigates the neighbor explosion problem of node-wise inference, it can handle 3-layer models on the large graphs (e.g., \textit{Friendster}). However, the results show that \sys still consistently outperforms node-wise inference. In particular, the speedup of \sys is several times for the small graphs (i.e., \textit{OGBN-Products} and \textit{Livejournal1}) but becomes over 10x for the large graphs. This is because \sys only accesses 1-hop neighbors for each node and thus still achieves better input node and computation sharing than node-wise inference.  

%can also share input nodes and computation under sampling inference, and large graphs makes it more difficult for node-wise inference to share input.

\begin{table}[t]
	\centering
	\fontsize{8}{10}\selectfont    %{字体尺寸}{行距}
	% \footnotesize
	\caption{Running time (in seconds) of sampling-based inference for 3-layer GCN. We use aberrations for the 4 datasets, i.e., \textit{OGBN-Products}, \textit{Livejournal1}, \textit{OGBN-Papers100M}, and \textit{Friendster}.}
	\label{tab:sampled}
	\begin{tabular}{c|cccc}
		%\toprule
		%  \cr
		\cmidrule(lr){1-5}
		& Prodts & Livej1 & Papers & Friend \cr
		\cmidrule(lr){1-5}
		Node-wise (16GB V100) & 17.7 & 30.1 & 1390 & 3530  \cr
		\sys(16GB V100) & 4.97 & 7.92 & 341 & 166  \cr
		Node-wise (32GB V100) & 11.4 & 23.3 & 1149 & 1895  \cr
		\sys(32GB V100) & 3.13 & 5.43 & 166 & 121   \cr
		\bottomrule
	\end{tabular}\vspace{0cm}
\trim
\end{table}     

% Although the neighbor expansion problem is much mitigated, \sys still outperform node-wise by up to 21x. By using sampled neighbors, we can avoid the problem brought by nodes with super large in-degrees in the graph. For sparse graphs like \textit{OGBN-Papers100M}, although the average in-degrees is 15.5, it still contains a large number of nodes that neighbor less than 10. Sampled neighbors can reduce a great deal of computation cost, the node-wise performance of using 16GB V100 and 32GB 100 are similar. However \sys can take the benefit of large GPU memory, which \sys runs twice as fast with 32GB V100 as with 16GB V100. The shorter inference time of \sys is attributed to node reorder and dynamic batch size control, which we will analyze in more detail. Both node-wise and layer-wise inference performs better as GPU memory size increases because of the use of a larger batch size, which leads to less data transfer from CPU to GPU. Due to the obvious inefficiencies of node-wise inference, we exclude it from the subsequent experiments.

%% file: sections/6_subsections/3_optimizations.tex
\begin{figure}[htbp]
	\centering
\includegraphics[width=\linewidth,scale=0.9]{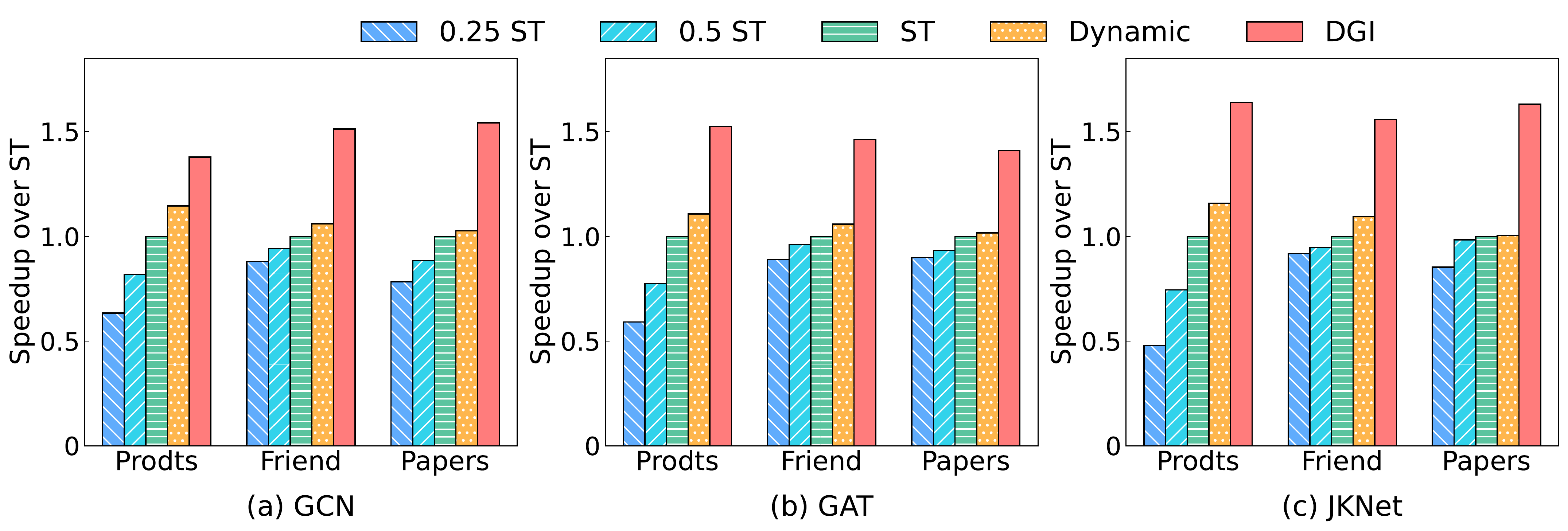}
\caption{Effect of dynamic batch size and node ordering for 3-layer models on V100-16GB GPU. Inference time is reported as speedup over \textit{ST}.
%We report the inference speedup over ST baseline Comparison between the inference execution time of \sys and the baselines. As the execution time for different graphs varies in scale, we normalize the execution time w.r.t. ST. The numbers under each dataset are the inference execution time of \sys (in seconds).
}
\label{fig:ablation_study}
\trim
\end{figure}

\subsection{Micro Experiments}\label{subsec:exp micro}

%. \textit{ST} represents an ideal case while \textit{0.5ST} and \textit{0.25ST}
\sys comes with two key optimizations, i.e., node reordering and dynamic batch size control. To evaluate their effects, we design 4 baselines: \textit{0.25ST}, \textit{0.5ST}, \textit{ST}, and \textit{Dynamic}, which all adopt the layer-wise inference scheme. \textit{ST} uses the best static batch size (found by grid search) that minimizes inference time. We also include \textit{0.5ST} and \textit{0.25ST}, which use 0.5 and 0.25 of the \textit{ST} batch size, respectively, and reflect practical scenarios where users choose a safe batch size.
\textit{Dynamic} uses dynamic batch size control but disables node reordering. We compare these baselines with DGI, which enables all optimizations.

\begin{figure}[t]
	\centering
\includegraphics[width=\linewidth,scale=0.9]{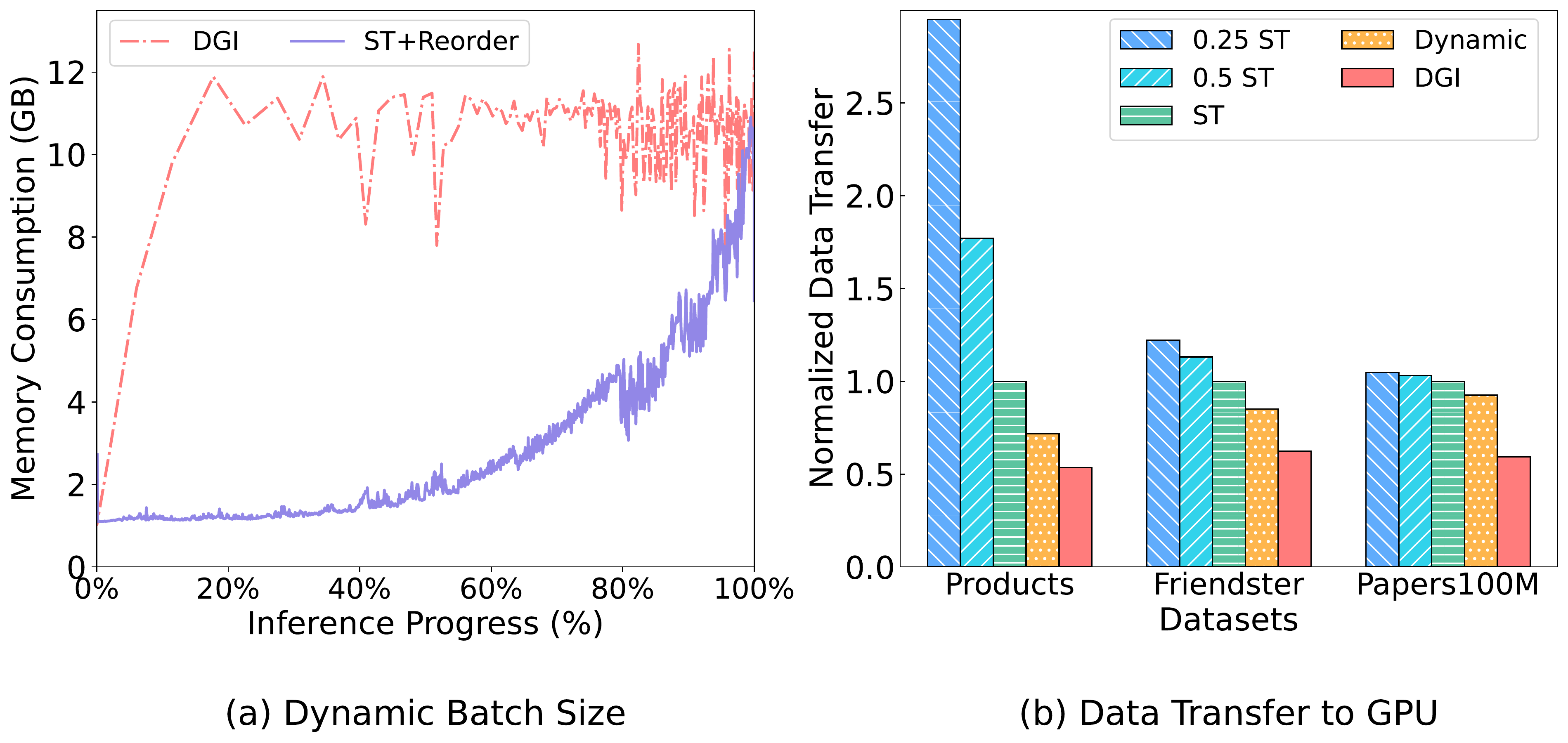}
\caption{The effects of node reordering and dynamic batch size control. We use 3-layer GCN and the \textit{Friendster} graph.}
\label{fig:micro-all}
\trim
\end{figure}

\begin{figure}[t]
	\centering
\includegraphics[width=\linewidth,scale=0.9]{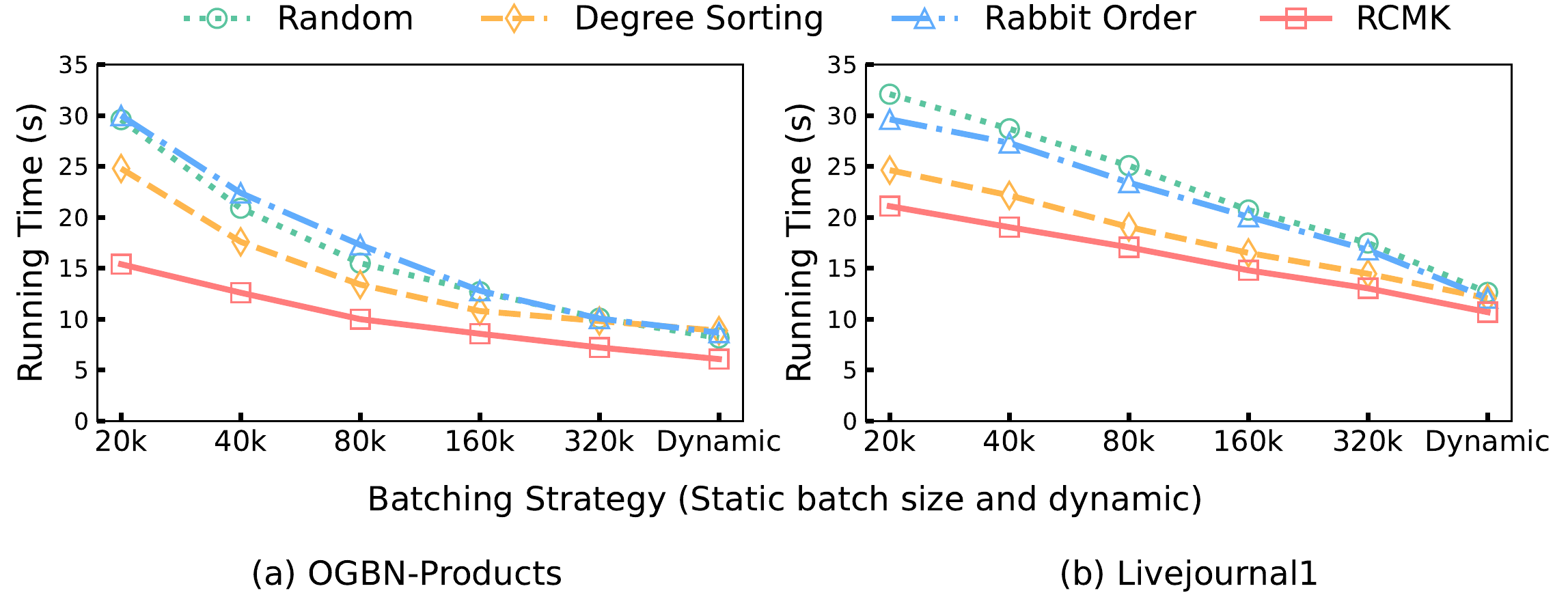}
\caption{\sys using different node reordering methods. The experiments are conduct using 3-layer GCN on V100-16GB GPU.}
\label{fig:reorder}
\trim
\end{figure}

%while the results on V100-32GB are similar and included in the supplementary material
% on V100-16GB GPU
Figure~\ref{fig:ablation_study} reports the inference time of \sys and the baselines. The results show that \textit{Dynamic} consistently outperforms \textit{ST}, which indicates that dynamic batch size control is effective. To understand its benefit, we plot the GPU memory consumption for \sys and \sys without dynamic batch size control (denoted as \textit{ST+Reorder}) in Figure~\ref{fig:micro-all}(a).
%We can make several observations from Figure~\ref{fig:ablation_study}. First, dynamic batch size control (DBSC) is effective as Dynamic often outperforms ST, which uses an ideal static batch size. To further explore the effect of DBSC, in Figure~\ref{fig:micro-all}(a), we compare the memory utilization of \sys and \textit{Reorder} (which disables DBSC and uses a static batch size).
The results show that \sys fully utilizes GPU memory except for the initial bootstrap period. In contrast, static batch size (i.e., \textit{ST+Reorder}) is limited by the memory consumption of the high degree nodes (in the last few batches) and thus has severe memory under-utilization. Figure~\ref{fig:ablation_study} also shows that node reordering gives another performance boost to \sys (by comparing \textit{Dynamic} with \sys). This is because node reordering organizes nodes with similar neighbors into the same batch to reduce the number of input neighbor embeddings. Figure~\ref{fig:micro-all}(b) reports the data transfer for the baselines. The results show that node reordering enables DGI to save up to 39.7\% of the data traffic compared with Dynamic.

There are many methods to reorder nodes in a graph to improve cache locality, which share the same spirit with the RCMK-based node recording in \sys. We compare RCMK with 3 alternatives: (i) \textit{Random}, which randomly shuffles the nodes; (ii) \textit{Degree Sorting}, which sorts the nodes by their in-degrees; (iii) Rabbit Order~\citep{rabbit}, which uses hierarchical community detection to determine node order. Figure~\ref{fig:reorder} shows that RCMK consistently yields shorter inference time than the other methods, which justifies using RCMK in \sys. This is because RCMK uses BFS to assign node IDs, and nodes with adjacent IDs are likely to share input nodes. We conjecture that Rabbit Order has poor performance because focus on improve graph locality by storing neighbors in CPU caches (typically tens of MB) for graph algorithms like PageRank. Thus, it generates small neighborhood and is not suitable for GNN inference.

%, and thus it is flexible to organize the nodes into batches

\stitle{Hardware choices} Cloud providers (e.g., AWS, Azure and Alibaba) offer machine instances with different memory capacities and GPUs, and DGI can adapt to different instances by storing data in either SSD or main memory. Thus, an interesting question is how to select a proper machine instance to conduct GNN inference. We report such results in Table~\ref{tab:cost and time}, which show that inference time significantly reduces when all data can be stored in main memory. Even if data does not fit in main memory, larger memory capacity still reduces inference time. Moreover, larger GPU memory is effective in reducing inference time. As advanced GPUs are very expensive on cloud, we suggest to ensure that the instance has enough main memory in the first place.

% \xiao{can we have more interesting configurations? small main memory+large GPU memory; large GPU memory+poor computation}   

%, which lead to several observations. First, inference time significantly reduces when switching data storage from SSD to main memory. Second, as shown previously, the benefit of using GPU with larger memory is more obvious for denser graphs (e.g., Friendster) than sparser graphs (e.g., Papers100M). Third, even for the same graph, the benefit of using larger GPU varies for different models. Take Papers100M for example, the JKNet has larger speedup than GCN and GAT when switching from M3 to M4. This is because JKNet requires more input than GCN and GAT in the last layer, and thus benefits more from the large batch size enabled by 32GB GPU memory.

\begin{table}[htbp]
%\fontsize{7}{9}\selectfont    %{字体尺寸}{行距}
% 	\footnotesize
\trim
\caption{Execution time of \sys (in seconds) on different machines. The models have 3 layers, and cases that store data on SSD are marked in bold. The first number is the capacity of main memory.}
\label{tab:cost and time}
\centering
\scalebox{0.8}{
	\begin{tabular}{c|ccc|ccc}
		%\toprule
		%  \cr
		\cmidrule(lr){1-7}
		\multirow{2}{*}{Machine }&
		\multicolumn{3}{c}{\textbf{OGBN-Papers100M}}&\multicolumn{3}{c}{\textbf{Friendster}} \cr
		\cmidrule(lr){2-7}
		& GCN & GAT & JKNet & GCN & GAT & JKNet \cr
		\cmidrule(lr){1-7}
		 128 GB + 16 GB T4 & \textbf{1828} & \textbf{2840} & \textbf{2030} & \textbf{3070} & \textbf{4510} & \textbf{3220} \cr
		192 GB + 16 GB T4 & \textbf{1590} & \textbf{2360} & \textbf{1730} &  \textbf{2660} & \textbf{3930} & \textbf{3030} \cr
		 488 GB + 16 GB V100 & 509 & 1074 & 659 & 582 & 1290 & 801 \cr
		1.5 TB + 32 GB V100 & 277 & 644 & 325 & 356 & 822 & 459    \cr
		\bottomrule
	\end{tabular}
}
%}
\trim
\end{table}

%% file: sections/6_related_works.tex
\section{Related Works}
\label{sec:related}

\stitle{Systems for GNN training} DGL~\citep{dgl} and PyTorch Geometric (PyG)~\citep{pyg} are state-of-the-art frameworks for GNN training. PyTorch-Direct~\citep{pytorch-direct} and TorchQuiver~\citep{torchquiver} train GNN on large graphs that exceed GPU memory by storing data on CPU memory and accessing them via CUDA unified virtual addressing (UVA). DistDGL~\citep{distdgl}, Neugraph~\citep{neugraph}, Roc~\citep{roc} and AliGraph~\citep{aligraph} scale GNN training to multiple machines. These and many other systems~\citep{GNNAdvisor, G3, Dorylus, Bgl, fuseGNN} are primarily designed for GNN training instead of inference. 

%In particular, training usually computes output embedding for a small number nodes in each iteration and conducts neighbor sampling, which alleviates the neighbor explosion problem of the node-wise inference. In contrast, inference usually uses the full neighbor set for accuracy and may compute embedding for all graph nodes, which makes node-wise inference inefficient and necessitates our layer-wise approach.       

%These systems mainly focus on GNN training, which do not provide easy and efficient support for inference.

\stitle{Computation optimizations for GNN} GCNP~\citep{channelpruning} computes node embedding efficiently by reducing the dimension of the intermediate embeddings. GNNAutoScale~\citep{GNNautoscale} uses the historical embeddings (i.e., computed in previous iterations) of neighbors to conduct aggregation. The two works trade accuracy for efficiency and are orthogonal to our \sys.
PCGraph~\citep{pcgraph} reduces data transfer overhead by caching node embeddings in GPU. Partition-based sparse matrix multiplication (SPMM) is used to run each layer of GNN models efficiently on FPGA~\cite{hardware}. These optimizations (i.e., embedding caching and layer execution) can also be integrated into \sys. To our knowledge, the idea of layer-wise inference is also seen in two code snippets~\cite{rgcncode, sagecode}, which however are manually written to handle simple GNN models with a linear structure only. We started the \sys project independently and our work formalizes the procedure of layer-wise inference, analyzes its advantages over node-wise inference, and we also design a system to make layer-wise inference general for GNN models and easy to use by translating the training code.

%DGL~\citep{dgl} and Exact~\citep{exact} snippet applies the idea of layer-wise inference for simple GNN models like GCN in its implementation. \sys can provide easy yet efficient GNN inference for models with different structures. 

%tried to accelerate GCN inference by using partition-based SPMM on hardware. Layer-wise in \sys is actually a row-based partition strategy and using DGL SPMM kernel. Partition-based is quite difficult to apply on more complex models -- need users hand-write kernel for each convolution while \sys are very easy to deploy. 

%\stitle{Graph processing systems}
%The graph processing system has been extensively studied for its strong ability to process data on graphs. Galois~\citep{galois} and Ligra~\citep{ligra} are well-known graph processing systems work on CPU. Cusha~\citep{cusha} and Gunrock~\citep{gunrock} speed up processing by using on faster computing machine GPU. GraphChi~\citep{graphchi} and GraphEne~\citep{graphene} support large-scale graph by using SSD.

Currently, we assume that \sys works for a static graph. This scenario is observed in many applications that periodically update the graph structure, GNN model and node embedding~\cite{pinterest,taobao}. A challenging scenario is to update node embedding for dynamic graphs in real-time. In particular, the graph is receiving updates (e.g., edge insertion/deletion, node feature changes), node embedding needs to be updated accordingly for  applications with high requirements on accuracy (e.g., recommendation and fraud detection). However, for an $L$-layer GNN model, a node/edge update can affect the embedding of its $L$-hop neighbors, which results in an enormous update set. Theory/algorithm may be required to limit the update set (e.g., skip a target node if changes to its embedding is smaller than a threshold in terms of Euclidean norm), and \sys can take the pruned update set as target nodes.      
 
%\sys works on static graph, which the graph are not changing during GNN training and inference. The graph structure and node embeddings will only update periodically rather than just-in-time. This scenario is widely used in industry. For example, PInterest~\citep{pinterest} outdated embedding generated in an offline process using the MapReduce framework. Taobao~\citep{taobao} runs GNN-based malicious account detection on a daily basis, rather than immediately after one transaction pops up.

%% file: sections/7_conclusion.tex
\section{Conclusions}
\label{sec:conclusions}

We build \sys, a system that makes inference easy and efficient for GNN models. We observe that the popular node-wise inference approach is inefficient and present a layer-wise inference approach, which is efficient but difficult to program. To mitigate such programming difficulty, \sys automatically traces the computation graph for GNN models, partitions the computation graph for execution, and manages node batching. \sys is general for different GNN models and inference requests, and supports out-of-core execution for large graphs. 
%We show that \sys is easy to use and much more efficient than node-wise inference.         

%To mitigate the programming difficult of layer-wise inference,  

%In this paper, we present \sys, an extension for the famous DGL system that atomically conducts GNN inference for large graphs without user effort. We observe that the existing node-wise inference scheme suffers from the neighbor explosion problem, and formalize a layer-wise inference scheme to avoid its deficiencies. \sys executes layer-wise inference automatically by tracing the computation graph of the GNN model and splitting the computation graph into layers using tailored rules. To organize the nodes in the graph into batches for inference, \sys adopts dynamic batch size control to free users from batch size tuning and node reordering to reduce the cost of transferring data to the GPU. With the use of SSD, machines with small CPU memory can scale to GNN on large graphs. Experiment results show that \sys can outperform node-wise inference by orders of magnitude and its optimizations are effective in reducing inference time. 